%% file: main.tex
\documentclass[10pt]{article} %
\usepackage[table]{xcolor}
\usepackage{xcolor}
\definecolor{bestcolor}{RGB}{252, 229, 205}
\newcommand{\best}[1]{\cellcolor{bestcolor}{\textbf{#1}}}

\usepackage[accepted]{tmlr}
\def\preprint{1}

\input{math_commands.tex}

\usepackage[pagebackref=true,breaklinks=true,colorlinks,linkcolor=blue,citecolor=blue,bookmarks=false]{hyperref}
\usepackage{url}
\usepackage{graphicx}
\usepackage{amsmath}
\usepackage{amssymb}

\let\emptyset\varnothing

\usepackage{booktabs}
\usepackage{multirow}

\usepackage{xspace}
\usepackage{enumitem}
\usepackage{caption}
\usepackage{subcaption}
\usepackage{wrapfig}

\usepackage[capitalize]{cleveref}
\crefformat{footnote}{#2\footnotemark[#1]#3}

\newcommand{\eg}{\textit{e.g.}\xspace}

\newcommand{\ie}{\textit{i.e.}\xspace}

\newcommand{\oursfull}{Composed Image Retrieval with Latent Diffusion (\ours)\xspace}
\newcommand{\ours}{\mbox{CompoDiff}\xspace}
\newcommand{\ourdataset}{SynthTriplets18M\xspace}

\newcommand{\rawrefimg}{\ensuremath{x_{i_R}}\xspace}
\newcommand{\rawcond}{\ensuremath{x_{c}}\xspace}
\newcommand{\rawcondtext}{\ensuremath{x_{c_T}}\xspace}
\newcommand{\rawcondnegtext}{\ensuremath{x_{c_{T^{\text{-}}}}}\xspace}
\newcommand{\rawcondmask}{\ensuremath{x_{c_M}}\xspace}
\newcommand{\rawtarimg}{\ensuremath{x_i}\xspace}
\newcommand{\rawtriplet}{\ensuremath{\langle \rawrefimg, \rawcond, \rawtarimg \rangle}\xspace}
\newcommand{\rawreftxt}{\ensuremath{x_{t_R}}\xspace}
\newcommand{\rawtartxt}{\ensuremath{x_t}\xspace}
\newcommand{\rawcaptriplet}{\ensuremath{\langle \rawreftxt, \rawcond, \rawtartxt \rangle}\xspace}
\newcommand{\rawcappair}{\ensuremath{\langle \rawreftxt, \rawtartxt \rangle}\xspace}

\newcommand{\refimg}{\ensuremath{z_{i_R}}\xspace}
\newcommand{\cond}{\ensuremath{z_{c}}\xspace}
\newcommand{\condtext}{\ensuremath{z_{c_T}}\xspace}

\newcommand{\tarimg}{\ensuremath{z_i}\xspace}

\title{CompoDiff: Versatile Composed Image Retrieval\\With Latent Diffusion}

\author{\small Geonmo Gu$^{*,\,1}$, Sanghyuk Chun$^{*,\,2}$, Wonjae Kim$^{2}$, HeeJae Jun$^{1}$, Yoohoon Kang$^{1}$, Sangdoo Yun$^{2}$\\
\,\\
\textnormal{$^{1}${NAVER Vision} \qquad $^{2}${NAVER AI Lab} \qquad $^*$ Equal contribution}
}

\def\month{07}  %
\def\year{2024} %
\def\openreview{\url{https://openreview.net/forum?id=mKtlzW0bWc}} %

\begin{document}

\maketitle

\begin{abstract}
This paper proposes a novel diffusion-based model, CompoDiff, for solving zero-shot Composed Image Retrieval (ZS-CIR) with latent diffusion. This paper also introduces a new synthetic dataset, named SynthTriplets18M, with 18.8 million reference images, conditions, and corresponding target image triplets to train CIR models. CompoDiff and SynthTriplets18M tackle the shortages of the previous CIR approaches, such as poor generalizability due to the small dataset scale and the limited types of conditions. CompoDiff not only achieves a new state-of-the-art on four ZS-CIR benchmarks, including FashionIQ, CIRR, CIRCO, and GeneCIS, but also enables a more versatile and controllable CIR by accepting various conditions, such as negative text, and image mask conditions. CompoDiff also shows the controllability of the condition strength between text and image queries and the trade-off between inference speed and performance, which are unavailable with existing CIR methods.
\ifx\preprint\undefined
The code and dataset samples are available at \texttt{Supplementary Materials}.
\else
The code and dataset are available at \url{https://github.com/navervision/CompoDiff}.
\fi
\end{abstract}

\section{Introduction}

Imagine a customer seeking a captivating cloth serendipitously found on social media but not the most appealing materials and colors. In this scenario, the customer needs a search engine that can process composed queries, \eg, the reference garment image along with text specifying the preferred material and color. This task has been recently formulated as \textit{Composed Image Retrieval (CIR)}. CIR systems offer the benefits of searching for visually similar items while providing a high degree of freedom to depict text queries as text-to-image retrieval. CIR can also improve the search quality by iteratively taking user feedback. 

\begin{figure}[t]
    \centering
    \includegraphics[width=\linewidth]{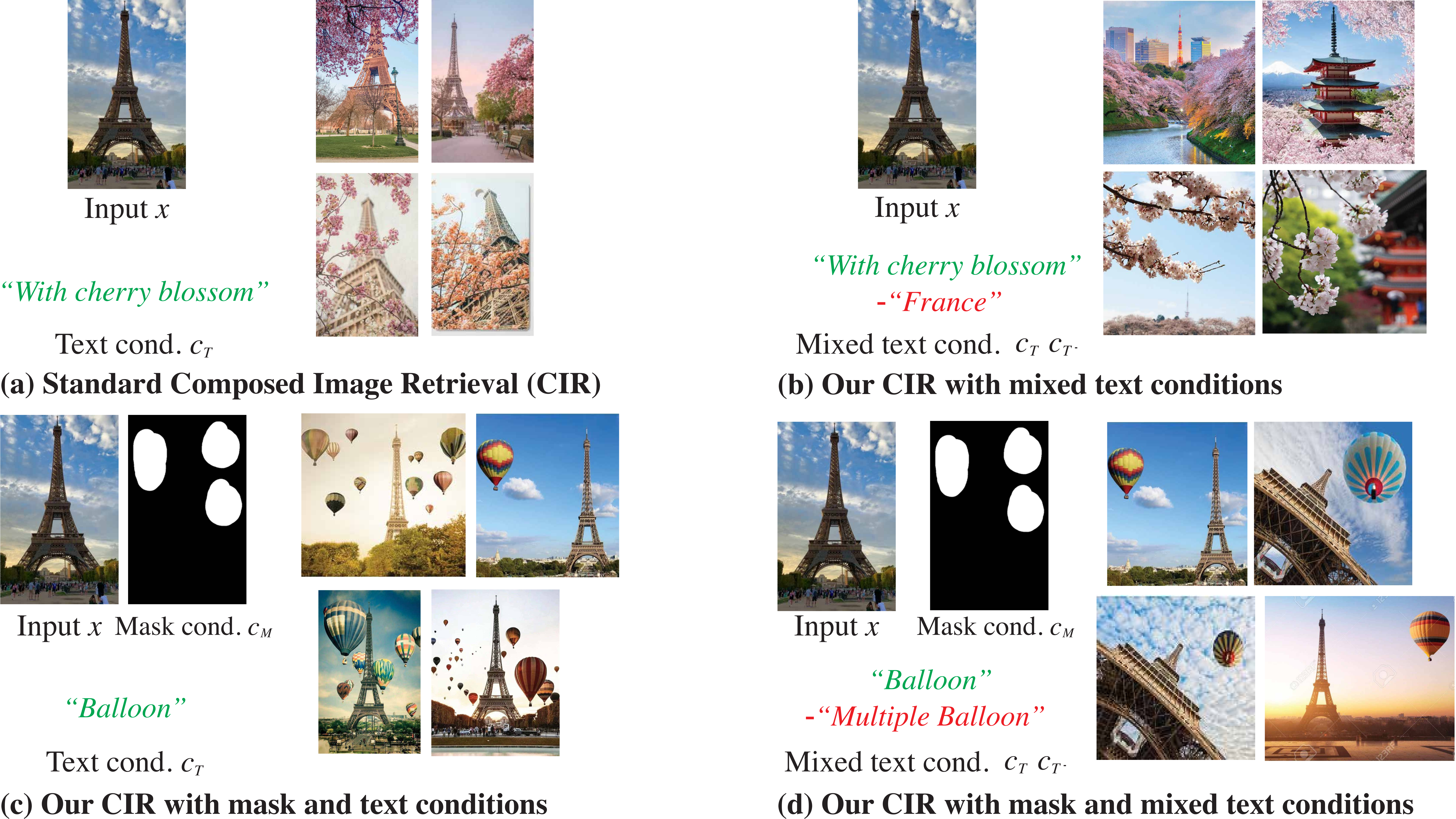}
    \caption{\small \textbf{Composed Image Retrieval (CIR) scenarios.} (a) A standard CIR scenario. (b-d) Our versatile CIR scenarios with mixed conditions (\eg, negative text and mask). Results by \ours on LAION-2B.}
    \label{fig:teaser}
    \vspace{-.5em}
\end{figure}

The existing CIR methods address the problem by combining image and text features using additional fusion models, \eg, $z_i = \texttt{fusion}(z_{i_R}, z_c)$ where $z_{i}$, $z_c$, $z_{i_R}$ are the target image, conditioning text, and reference image features, respectively\footnote{Throughout this paper, we will use $x$ to denote raw data and $z$ to denote a vector encoded from $x$.}.
Although the fusion methods have shown great success,
they have fundamental limitations. First, the fusion module is not flexible; it cannot handle versatile conditions beyond a limited textual one. 
For instance, a user might want to include a negative text that is not desired for the search (\rawcondnegtext) (\eg, an image $+$ ``with cherry blossom'' $-$ ``France'', as in \cref{fig:teaser} (b)), indicate where (\rawcondmask) the condition is applied (\eg, an image $+$ ``balloon'' $+$ indicator, as in \cref{fig:teaser} (c)), or construct a complex condition with a mixture of them.
Furthermore, once the fusion model is trained, it will always produce the same $z_i$ for the given $z_{i_R}$ and $z_c$ to users. However, a practical retrieval system needs to control the strength of conditions by its applications or control the level of serendipity.
Second, they need a pre-collected human-verified dataset of triplets \rawtriplet consisting of a reference image (\rawrefimg), a text condition (\rawcond), and the corresponding target image (\rawtarimg).
However, obtaining such triplets is costly and sometimes impossible; therefore, the existing CIR datasets are small-scale (\eg, 30K triplets for Fashion-IQ \citep{fashioniq} and 36K triplets for CIRR \citep{cirr}), resulting in a lack of generalizability to other datasets. 

We aim to achieve a generalizable CIR model with diverse and versatile conditions by using latent diffusion. We treat the CIR task as a conditional image editing task on the latent space, \ie, $z_i = \texttt{Edit}(z_{i_R} | z_c, \ldots)$. Our diffusion-based CIR model, named \ours, can easily deal with versatile and complex conditions, benefiting from the flexibility of the latent diffusion model \citep{rombach2022latentdiffusion} and the classifier-free guidance \citep{ho2022classifier}. Note that the purpose of image editing tasks and CIR tasks are different; an image editing model aims to find any image satisfying the changes due to the instruction, while a CIR model aims to find a generalized concept of the given composed queries. For example, assume a picture of a dog on a grass field and an instruction ``change the dog to a cat''. Here, even if we assume a perfect image generator can generate an image of a cat on a grass field, it could not achieve a good retrieval performance because it cannot generate all possible cat images in the database. In other words, the purpose of image editing is ``precision'', while the purpose of image retrieval is ``recall''; it makes the difference between image editing and retrieval. In our experiments, we support this claim by showing directly using an editing model performs much worse than other CIR methods. Therefore, we need a specialized model for CIR rather than directly using the image editing model. In this paper, we tackle the problem by using a latent diffusion model instead of an image-level diffusion model and introducing various retrieval-specific conditions, such as mask conditions. We train a latent diffusion model that translates the embedding of the reference image (\refimg) into the embedding of the target image (\tarimg) guided by the embedding of the given text condition (\cond). As shown in \cref{fig:teaser}, \ours can handle various conditions, which is not possible with the standard CIR scenario with the limited text condition \rawcondtext. Although our method has an advantage over the existing fusion-based CIR methods in terms of versatility, \ours also needs to be trained with triplet datasets where the scale of the existing CIR datasets is extremely small.

We address the dataset scale issue by synthesizing a vast set of high-quality \textbf{18.8M} triplets of \rawtriplet. Our approach is fully automated without human verification; hence, it is scalable even to 18.8M. We follow InstructPix2Pix (IP2P) \citep{brooks2022instructpix2pix} for synthesizing triplets, while our dataset contains $\times$40 more triplets and $\times$12.5 more keywords (\eg, objects, background details, or textures) than IP2P. Our massive dataset, named \textbf{\ourdataset}, is over 500 times larger than existing CIR datasets and covers a diverse and extensive range of conditioning cases, resulting in a notable performance improvement for any CIR model. For example, ARTEMIS \citep{delmas2022artemis} trained exclusively with \ourdataset shows outperforming zero-shot performance even than its FashionIQ-trained counterpart (40.6 vs. 38.2).

To show the generalizability of the models, we evaluate the models on the ``zero-shot'' (ZS) CIR scenario using four CIR benchmarks: FashionIQ \citep{fashioniq}, CIRR \citep{cirr}, CIRCO \citep{circo}, and GeneCIS \citep{genecis}; \ie, we report the retrieval results by the models trained on our \ourdataset and a large-scale image-text paired dataset without access to the target triplet datasets. In all experiments, \ours achieves the best zero-shot performances with significant gaps (See \cref{tab:main}). Moreover, we observe that the fusion-based approaches solely trained on \ourdataset (\eg, Combiner \citep{baldrati2022clip4cir}) show comparable or outperforming zero-shot CIR performances compared to the previous SOTA zero-shot CIR methods, \eg, Pic2Word \citep{saito2023pic2word}, SEARLE \citep{circo} and LinCIR \citep{lincir}. Furthermore, we qualitatively observe that the retrieval results of \ours are semantically better than previous zero-shot CIR methods, such as Pic2Word, on a large-scale image database, \eg, LAION-2B \citep{schuhmann2022laion}.

Another notable advantage of \ours is the controllability of various conditions during inference, which is inherited from the nature of diffusion models. Users can adjust the weight of conditions to make the model focus on their preference. Users can also manipulate randomness to vary the degree of serendipity. In addition, \ours can control the speed of inference with minimal sacrifice in retrieval performance, accomplished by adjusting the number of sampling steps in the diffusion model. As a result, \ours can be deployed in various scenarios with different computational budgets. All of these controllability features are achievable by controlling the inference parameters of classifier-free guidance without any model training.

Our contributions are first to show the effectiveness of diffusion models on non-generative tasks, such as multi-modal retrieval and data generation. More specifically: (1) We propose a diffusion-based CIR method, named \ours.
\ours can handle various conditions, such as the negative text condition, while the previous methods cannot. We also can control the strength of conditions, \eg, more focusing on the reference image while the changes by the text should be small. (2) We generate \ourdataset, a diverse and massive synthetic dataset of 18M triplets that can make CIR models achieve zero-shot (ZS) generalizability. Our data generation process is fully automated and easily scalable. (3) Our experiments support the effectiveness of \ours quantitatively (significant ZS-CIR performances on FashionIQ, CIRR, CIRCO, and GeneCIS datasets) and qualitatively (ZS-CIR on a billion-scale DB, or image decoding using unCLIP generator).

\section{Related Works}

\begin{figure}[t]
    \centering
    \includegraphics[width=\linewidth]{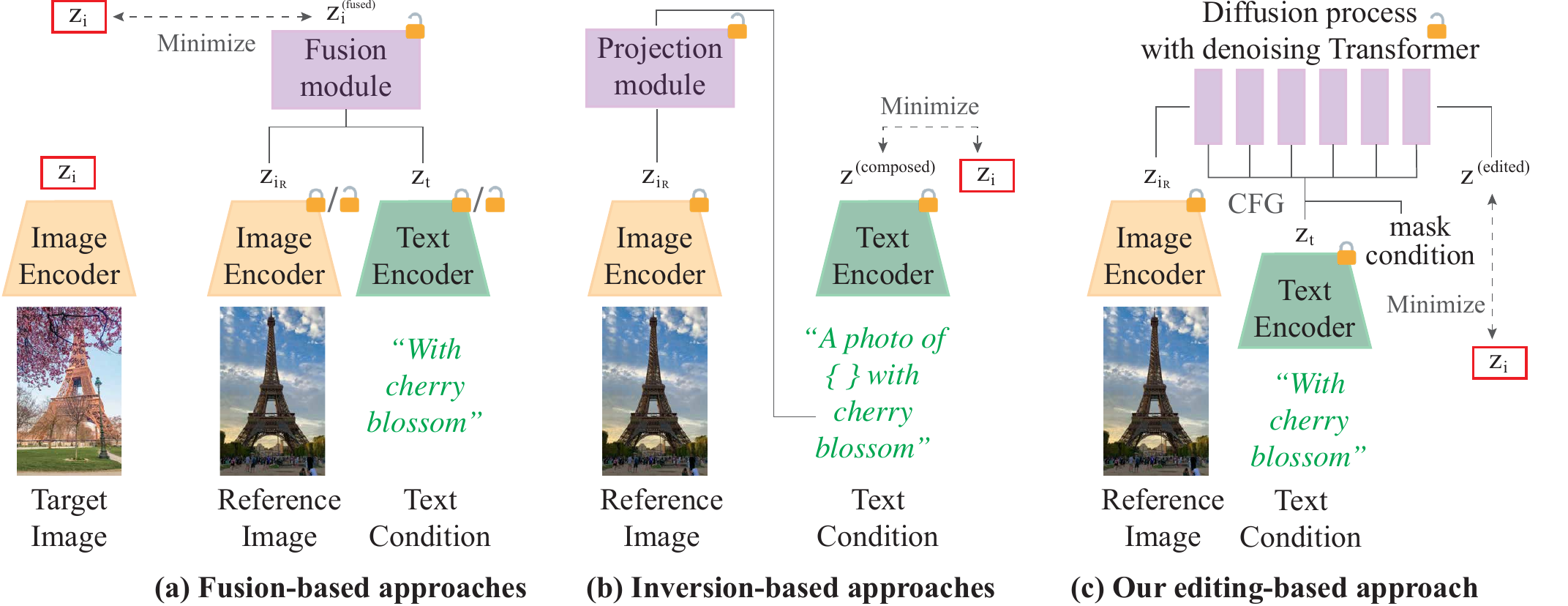}
    \caption{\small {\bf Comparisons of CIR methods.} (a) Fusion-based methods (\eg, ARTEMIS \citep{delmas2022artemis} and Combiner \citep{baldrati2022clip4cir}) make a fused feature from image feature $z_{i_R}$ and text feature $z_c$. (b) Inversion-based methods (\eg, Pic2Word \citep{saito2023pic2word}, SEARLE \citep{circo} and LinCIR \citep{lincir}) project $z_{i_R}$ into the text space, then perform text-to-image retrieval. (c) We apply a diffusion process to $z_{i_R}$ with classifier-free guidance by additional conditions. (b) and (c) use frozen encoders, and (a) usually tunes the encoders.}
    \label{fig:compodiff_overview}
    \vspace{-.5em}
\end{figure}

\paragraph{Composed image retrieval.}
The mainstream CIR models have focused on \textit{multi-modal fusion methods}, which combine image and text features extracted from separate visual and text encoders, as in \cref{fig:compodiff_overview} (a). For example, \citet{vo2019tirg} and \citet{yu2020curlingnet} used CNN and RNN, and \citet{chen2022mur} and \citet{Anwaar_Labintcev_Kleinsteuber_2021} used CNN and Transformer. More recent methods address CIR tasks by leveraging external knowledge from models pre-trained on large-scale datasets such as CLIP \citep{radford2021clip}. For example, Combiner \citep{baldrati2022clip4cir} fine-tunes the CLIP text encoder on the triplet dataset to satisfy the relationship of $z_{i} = z_{i_R} + z_{c}$, and then trains a Combiner module on top of the encoders. However, these fusion methods still require expensive pre-collected and human-verified triplets of the target domain.

To solve this problem, recent studies aim to solve CIR tasks in a zero-shot manner. \textit{Inversion-based zero-shot CIR methods}, such as Pic2Word \citep{saito2023pic2word}, SEARLE \citep{circo} and LinCIR \citep{lincir}, tackle the problem through text-to-image retrieval tasks where the input image is projected into the condition text -- \cref{fig:compodiff_overview} (b). Note that our zero-shot CIR scenario is slightly different from theirs; our zero-shot CIR denotes that the models are not trained on the target triplet datasets, but trained on our synthetic dataset, \ourdataset, and image-text paired dataset, \eg, LAION-2B \citep{schuhmann2022laion}. Meanwhile, \citet{saito2023pic2word}, \citet{circo} and \citet{lincir} use the term zero-shot when the CIR models are trained without a triplet dataset. 

All the existing CIR models only focus on text conditions (\condtext) (\eg, \cref{fig:teaser} (a)) and have difficulties in handling versatile scenarios (\eg, \cref{fig:teaser} (b-d)) with a lack of the controllability. On the other hand, our method enables multiple various conditions and controllabilities with strong zero-shot performances by employing (1) a latent diffusion model \citep{rombach2022latentdiffusion} with classifier-free guidance \citep{ho2022classifier} and (2) a massive high-quality synthetic dataset, \ourdataset.

Concurrent with our work, CoVR \citep{ventura2024covr}, VDG \citep{jang2024visual}, and MagicLens \citep{zhang2024magiclens} consider constructing training triplets from the existing image datasets, namely, collecting image pairs and synthesizing the modification texts using LLMs. Although they showed empirically good performances, their approaches need real images where it is often very difficult to generate plausible modification instructions between two real images. Meanwhile, our synthetic dataset is more controllable than their datasets. Namely, we can exactly guide the visual difference between two images with the given instructions.

\paragraph{Dataset creation with diffusion models.}
A conventional data collection process for \rawtriplet is two-staged: collecting candidate reference-target image pairs and manually annotating the modification sentences by human annotators. For example, FashionIQ \citep{fashioniq} collects the candidate pairs from the same item category (\eg, shirt, dress, and top) and manually annotates the relative captions by crowd workers. CIRR \citep{cirr} gathers the candidate pairs from real-life images from NLVR$^2$ \citep{suhr2018corpus}. The triplet collection process inevitably becomes expensive, making it difficult to scale CIR datasets. We mitigate this problem by generating a massive synthetic dataset instead of relying on human annotators.

Recently, there have been attempts to generate synthetic data to improve model performance \citep{brooks2022instructpix2pix, Nair_Bandara_Patel_2022, Shipard_Wiliem_Thanh_Xiang_Fookes_2023} by exploiting the powerful generation capabilities of diffusion models. In particular, \citet{brooks2022instructpix2pix} proposes a generation process for \rawtriplet to train an image editing model. We scale up the dataset synthesis process of \citet{brooks2022instructpix2pix} from 1M triplets to 18.8M. Our generation process contains more diverse triplets by applying the object-level modification process, resulting in better CIR performances on the same 1M scale and easily scalable to a larger number of samples.

\section{\ours: Composed Image Retrieval with Latent Diffusion}
\label{sec:method}

This section introduces \oursfull
which employs a diffusion process in the frozen CLIP latent feature space with classifier-free guidance. Unlike previous latent diffusion models, we use a Transformer-based denoiser. We train our model with various tasks, such as text-to-image (T2I) generation, masked T2I and triplet-based generation to handle various conditions (\eg, negative text or mask), while the previous CIR methods only limit beyond the positive text instruction.

\subsection{Preliminary: Diffusion model}
\label{subsec:diffusion}
Given a data sample from the distribution ($x \sim p(x)$) diffusion model (DM) defines a Markov chain of latent variables $z_1, \ldots, z_T$ as: $q(z_t | z_{t-1}) = \mathcal N(z_t; \sqrt \alpha_t z_{t-1}, (1 - \alpha_t) I)$. It is proven that (1) the posterior $q_{z_{t-1} | z_t}$ is approximated to a Gaussian distribution with a diagonal covariance, and (2) if the size of chain $T$ is sufficiently large, then $z_T$ will be approximated to $\mathcal N(0, I)$. Namely, DMs are a general probabilistic method that maps an input distribution $p(x)$ to a normal distribution $\mathcal N(0, I)$, without any assumption on the input data $x$. From this observation, \citet{rombach2022latentdiffusion} proposed to learn DMs on a latent space (\ie, latent diffusion), which brings a huge improvement in efficiency.
In practice, we train a denoising module $\epsilon_\theta(z_t, t)$ (\ie, $z_{t-1} = \epsilon_\theta(z_t, t)$) where the timestamp $t$ is conditioned by employing time embeddings $e_t$.

The recent DMs use classifier-free guidance (CFG) \citep{ho2022classifier} for training conditional DMs without a pre-trained classifier \citep{dhariwal2021diffusion}.
In this paper, we provide various conditions via CFG to take two advantages.
First, we can easily control the intensity of the condition. Second, we can extend the conditions beyond a single text condition, \eg, a negative text condition or a mask condition.

\ours has two distinct differences from the previous latent diffusion approaches, such as StableDiffusion (SD) \citep{rombach2022latentdiffusion}. First, SD performs the diffusion process on 64 $\times$ 64-dimensional VQ-GAN \citep{vqgan} latent space. Meanwhile, the diffusion process of \ours is performed on the CLIP latent embedding space.
Therefore, the edited features by \ours can be directly used for retrieval on the CLIP space, while SD cannot.
The Dall-e2 prior \citep{dall-e2} performs the diffusion process on the CLIP embedding space, but \ours takes inputs and conditions differently, as described below.

As the second contribution, \ours uses a different architecture for the de-noising module. While SD and other DM models are based on U-Net structure \citep{unet}, \ours uses a simple Transformer \citep{vaswani2017attention}. Moreover, \ours is designed to handle multiple conditions, such as masked conditions, and a triplet relationship. SD cannot handle the localized condition and is designed for a pairwise relationship (\eg, text-to-image generation). \ours also handles the condition different from Dalle-2 prior. Dalle-2 prior handles conditions as the input of the diffusion model, but our \ours diffusion Transformer takes the conditions via the cross-attention mechanism.
This design choice makes the inference speed of \ours faster. If the conditions are concatenated to the input tokens, the inference speed will be highly degenerated \citep{song2022vidt}. \cref{tab:ablation_stage2} shows that the structure taking all conditions as the concatenated input (\eg, Dalle-2 prior-like) is three times slower than our cross-attention approach.

\begin{figure}[t]
    \centering
    \includegraphics[width=\linewidth]{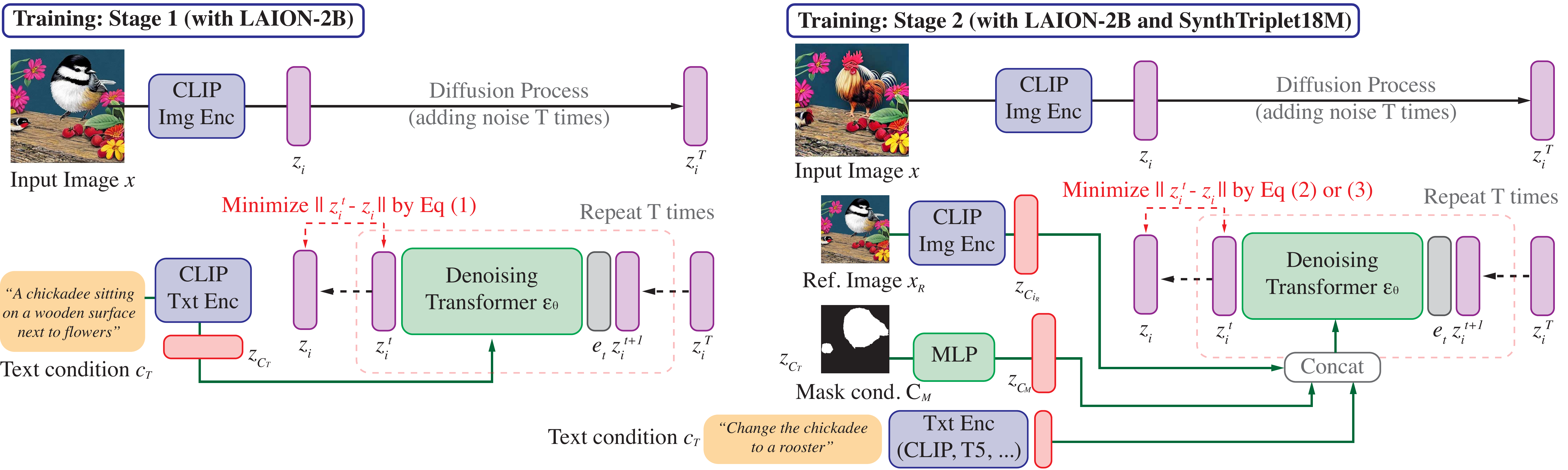}
    \caption{\small \textbf{Training overview.} Stage 1 is trained on LAION-2B with \cref{eq:stage1-t2i}. For stage 2, we alternatively update Denoising Transformer $\epsilon_\theta$ on LAION-2B with \cref{eq:stage1-t2i} and \ref{eq:stage2-inpaint} and \ourdataset with \cref{eq:stage2-triplet}.}
    \label{fig:method_training_overview}
    \vspace{-.5em}
\end{figure}
\subsection{Training}
\label{subsec:training}

\ours uses a two-stage training strategy as illustrated in \cref{fig:method_training_overview}. In stage 1, we train a text-to-image latent diffusion model on LAION-2B. In stage 2, we fine-tune the model on our synthetic triplet dataset, \ourdataset, and LAION-2B. Below, we describe the details of each stage.

In \textbf{stage 1}, we train a transformer decoder to convert CLIP textual embeddings into CLIP visual embeddings. This stage is similar to training the Dalle-2 prior, but our model takes only two tokens; a noised CLIP image embedding and a diffusion timestep embedding. The Dalle-2 prior model is computationally inefficient because it also takes 77 encoded CLIP text embeddings as an input. However, \ours uses the encoded text embeddings as conditions through cross-attention mechanisms, which speeds up the process by a factor of three while maintaining similar performance (See \cref{subsec:abl}). Instead of using the noise prediction of \citet{ho2020denoising}, we train the transformer decoder to predict the denoised $z_i$ directly due to the stability.

Now, we introduce the objective of the first stage with CLIP image embeddings of an input image $z_i$, encoded CLIP text embeddings for text condition $z_{c_T}$, and the denoising Transformer $\epsilon_\theta$:
\begin{equation}
\label{eq:stage1-t2i}
\mathcal{L}_\text{stage1} = \mathbb{E}_{t\sim[1,T]}  \|z_i - \epsilon_\theta(z_i^{(t)},t \vert z_{c_T}) \|^2
\end{equation}
During training, we randomly drop the text condition by replacing $z_{c_T}$ with a null text embedding $\emptyset_{c_T}$ in order to induce CFG. We use the empty text CLIP embedding (``'') for the null embedding.

In \textbf{stage 2}, we incorporate condition embeddings, injected by cross-attention, into CLIP text embeddings, along with CLIP reference image visual embeddings and mask embeddings (See \cref{fig:method_training_overview}).
We fine-tune the model with three different tasks: a conversion task that converts textual embeddings into visual embeddings (\cref{eq:stage1-t2i}), a mask-based conversion task (\cref{eq:stage2-inpaint}), and the triplet-based CIR task (\cref{eq:stage2-triplet}). The first two tasks are trained on LAION-2B and the last one is trained on \ourdataset.

\textbf{The mask-based conversion task} learns a diffusion process that recovers the full image embedding from a masked image embedding.
As we do not have mask annotations, we extract masks using a zero-shot text-conditioned segmentation model, CLIPSeg \citep{luddecke2022image}. We use the nouns of the given caption for the CLIPSeg conditions. Then, we add a Gaussian random noise to the mask region of the image and extract $z_{i, \text{masked}}$. We also introduce mask embedding $z_{c_M}$ by projecting a 64$\times$64 resized mask to the CLIP embedding dimension using a MLP. Now, the mask-based conversion task is defined as follows:
\begin{equation}
\label{eq:stage2-inpaint}
\mathcal{L}_\text{stage2\_masked\_conversion} = \mathbb{E}_{t\sim[1,T]}  \|z_i - \epsilon_\theta(z_{i, \text{masked}}^{(t)},t \vert z_{c_T}, z_{i, \text{masked}}, z_{c_M}) \|^2,
\end{equation}
Finally, we introduce the triplet-based training objective to solve CIR tasks on \ourdataset as follows:
\begin{equation}
\label{eq:stage2-triplet}
\mathcal{L}_\text{stage2\_triplet} = \mathbb{E}_{t\sim[1,T]}  \|z_{i_T} - \epsilon_\theta(z_{i_T}^{(t)},t \vert z_{c_T}, z_{i_R}, z_{c_M}) \|^2,
\end{equation}
where $z_{i_R}$ is a reference image feature and $z_{i_T}$ is a modified target image feature.

We update the model by randomly using one of \cref{eq:stage1-t2i}, \cref{eq:stage2-inpaint}, and \cref{eq:stage2-triplet} with the proportions 30\%, 30\%, 40\%.
As stage 1, the stage 2 conditions are randomly dropped except for the mask conditions. We use an all-zero mask condition for the tasks that do not use a mask condition. When we drop the image condition of \cref{eq:stage2-triplet}, we replace $z_{i_R}$ with the null image feature, an all zero vector.

Our two-staged training strategy is not a mandatory procedure, but it brings better performances. Conceptually, stage 1 is a pre-training of a diffusion model (DM) with pairwise image-text relationships. The stage 2 is for a fine-tuning process using triplet relationships. Lastly, the alternative optimization strategy for stage 2 is for helping optimization, not for resolving the instability. As shown in \cref{tab:ablation_stage2}, \ours can be trained only with \cref{eq:stage2-triplet}, but adding more objective functions makes the final performances stronger.
The two-staged training strategy is also widely used for the other CIR methods, such as SEARLE \citep{circo} or Combiner \citep{baldrati2022clip4cir}.
In terms of diffusion model fine-tuning, we argue that our method is not specifically complex than other fine-tuning methods, such as ControlNet \citep{controlnet}.

\begin{figure}[t]
    \centering
    \includegraphics[width=\linewidth]{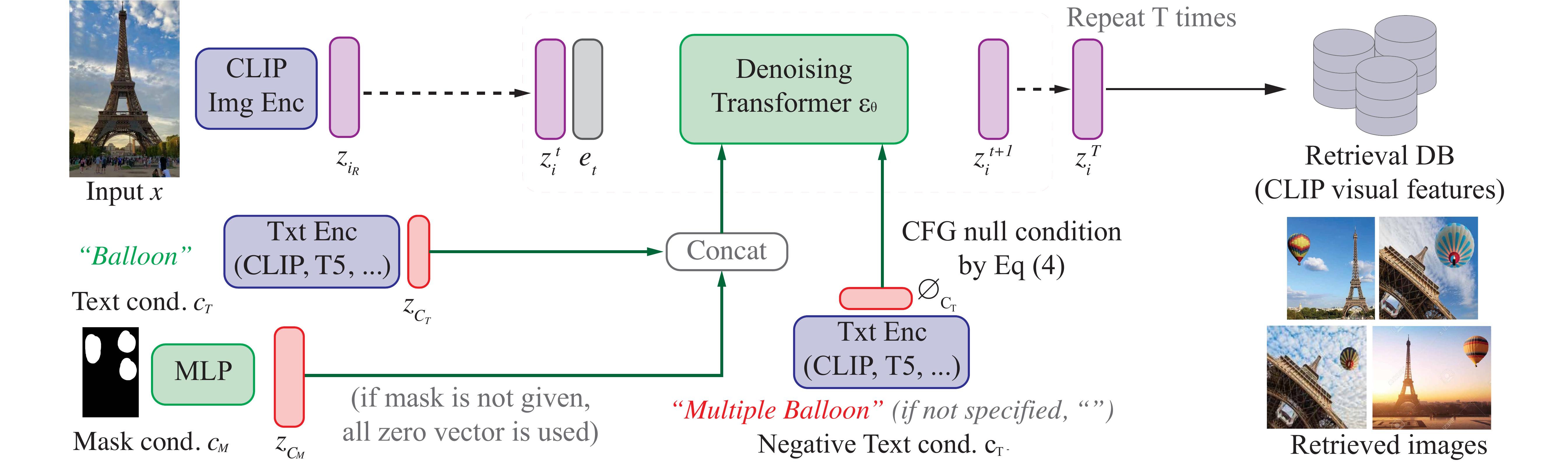}
    \caption{\small \textbf{Inference overview.} Using the denoising transformer $\varepsilon_\theta$ trained by Stage 1 and 2 (\cref{fig:method_training_overview}), we perform composed image retrieval (CIR). We use the classifier-free guidance by \cref{eq:cfg} to transform the input reference image to the target image feature, and perform image-to-image retrieval on the retrieval DB.}
    \label{fig:method_inference_overview}
    \vspace{-.5em}
\end{figure}
\subsection{Inference}
\label{subsec:inference}
\cref{fig:method_inference_overview} shows the overview. Given a reference image feature $z_{i_R}$, a text condition feature $z_{c_T}$, and a mask embedding $z_{c_M}$, we apply a denoising diffusion process based on CFG \citep{ho2022classifier} as follows:
\begin{equation}
\label{eq:cfg}
{
\begin{aligned}
\tilde{\epsilon}_\theta(z_{i}^{(t)},t \vert z_{c_T}, z_{i_R}, z_{c_M}) = \epsilon_\theta(z_{i}^{(t)},t \vert \emptyset_{c_T}, \emptyset_{i_R}, z_{c_M}) &+ w_I(\epsilon_\theta(z_{i}^{(t)},t \vert \emptyset_{c_T}, z_{i_R}, z_{c_M}) -\epsilon_\theta(z_{i}^{(t)},t \vert \emptyset_{c_T}, \emptyset_{i_R}, z_{c_M})) \\
&+ w_T(\epsilon_\theta(z_{i}^{(t)},t \vert z_{c_T}, z_{i_R}, z_{c_M}) -\epsilon_\theta(z_{i}^{(t)},t \vert \emptyset_{c_T}, z_{i_R}, z_{c_M}))
\end{aligned}
}
\end{equation}

where $\emptyset$ denotes null embeddings, \ie, the empty text (``'') CLIP textual embedding for the text null embedding and an all-zero vector for the image null embedding. $w_I$ and $w_T$ are weights for controlling the effect of image query or text query, respectively. We can control the degree of the visual similarity ($w_I$) or the text instruction ($w_T$) by simply adjusting the weight values without any training. One of the advantages of \cref{eq:cfg} is the ability to handle various conditions at the same time. When using negative text, we simply replace $\emptyset_{i_T}$ with the CLIP text embeddings $c_{T^-}$ for the negative text. These advantages are not available by the previous CIR methods without additional training.

\begin{wrapfigure}{r}{0.3\linewidth}
    \centering
    \vspace{-2.5em}
    \includegraphics[width=\linewidth]{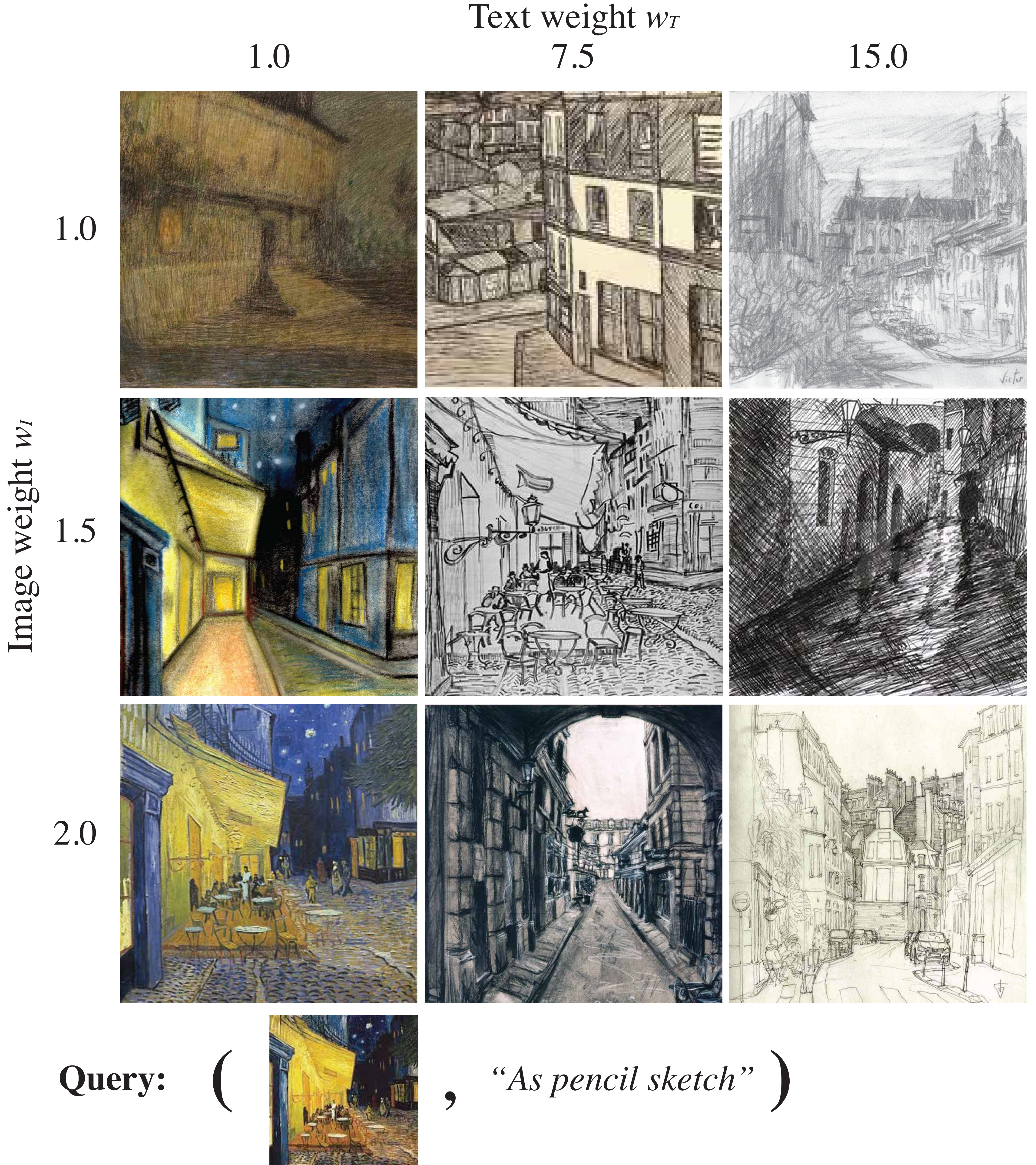}
    \vspace{-1.5em}
    \caption{\small Inference condition control by varying $w_I$, $w_T$ in \cref{eq:cfg}.}
    \label{fig:cfg_variation_images}
    \vspace{-3.5em}
\end{wrapfigure}
Another advantage of CFG is the controllability of the queries without training, \eg, it allows to control the degree of focus on image features to preserve the visual similarity with the reference by simply adjusting the weights $w_I$ or $w_T$.
We show the top-1 retrieved item by varying the image and text weights $w_I$ and $w_T$ from LAION-2B in \cref{fig:cfg_variation_images}. By increasing $w_I$, \ours behaves more like an image-to-image retrieval model. Increasing $w_T$, on the other hand, makes \ours focus more on the ``pencil sketch'' text condition. We use ($w_I$, $w_T$) = (1.5, 7.5) for our experiments. The full retrieval performance by varying $w_I$ and $w_T$ is shown in \cref{fig:heatmaps}.

As our model is based on a diffusion process, we can easily control the balance between the inference time and the retrieval quality of the modified feature by varying step size. In practice, we set the step size to 5 or 10, which shows the best trade-off.

\begin{figure}[t]
    \centering
    \includegraphics[width=.98\linewidth]{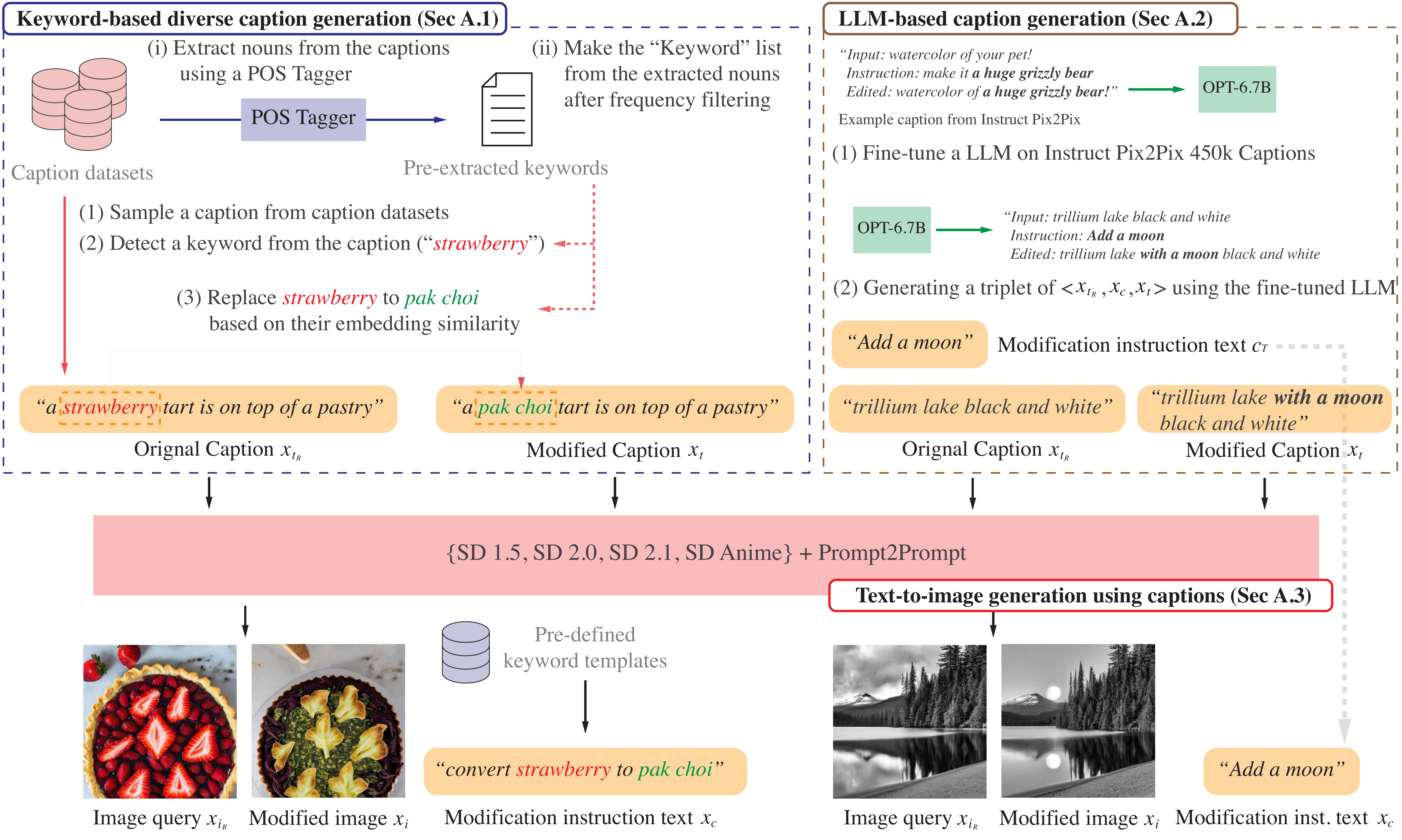}
    \caption{\small {\bf Overview of the generation process for \ourdataset.} \rawtriplet from \rawcaptriplet.
    }
    \label{fig:dataset_generation_overview}
\end{figure}

\section{\ourdataset: Massive High-Quality Synthesized Dataset}
\label{sec:cir_dataset_gen}

\begin{wraptable}{r}{0.5\linewidth}
\small
\centering
\setlength{\tabcolsep}{3pt}
\vspace{-1.5em}
\begin{tabular}{lcc}
\toprule
  & IP2P & \ourdataset \\ \midrule
\rawcaptriplet {\scriptsize (before filtering)} & 452k & 500M \\
\rawcaptriplet {\scriptsize (after filtering)}      & 313k & 60M \\
Unique object terms & 47,345 & \textbf{586,369} \\
\midrule
\rawtriplet ~~{\scriptsize (Keyword-based)} & - & 11.4M \\
\rawtriplet ~~{\scriptsize (LLM-based)} & 1M & 7.4M \\ \midrule
\rawtriplet ~~{\scriptsize (Total)}   & 1M & \textbf{18.8M} \\
\bottomrule
\end{tabular}
\caption{\small {\bf Dataset statistics.} 
\rawcaptriplet denotes the triplet of \textit{captions}, \ie, \{original caption, instruction, and modified caption\}, and \rawtriplet denotes the \textit{CIR triplet} of \{original image, instruction, and modified image\}.
}
\label{tab:keyword_stats}
\vspace{-3em}
\end{wraptable}
CIR requires a dataset of triplets \rawtriplet of a reference image (\rawrefimg), a condition (\rawcond), and the corresponding target image (\rawtarimg). Instead of collecting a dataset by humans, we propose to automatically generate massive triplets by using generative models. We follow the main idea of Instuct Pix2Pix (IP2P) \citep{brooks2022instructpix2pix}. First, we generate \rawcaptriplet where \rawreftxt is a reference caption, \rawcond is a modification instruction text, and \rawtartxt is the caption modified by \rawcond. We use two strategies to generate \rawcaptriplet: (1) We collect massive captions from the existing caption datasets and generate the modified captions by replacing the keywords in the reference caption (\cref{subsec:gen_keyword_supp}). (2) We fine-tune a large language model, OPT-6.7B \citep{zhang2022opt}, on the generated caption triplets from \citet{brooks2022instructpix2pix}
(\cref{subsec:gen_llm_supp}). After generating massive triplets of \rawcaptriplet, we generate images from the caption triplets using StableDiffusion (SD) and Prompt-to-Prompt \citet{hertz2022prompttoprompt} following IP2P (\cref{subsec:triplet_gen_supp}). We employ CLIP-based filtering to ensure high-quality triplets (\cref{subsec:filtering_supp}). The entire generation process is illustrated in \cref{fig:dataset_generation_overview}.

Compared to manual dataset collections \citep{fashioniq, cirr}, our approach can easily generate more diverse triplets even if a triplet rarely occurs in reality (See the examples in \cref{fig:dataset_generation_overview}). For example, FashionIQ only contains text instructions for fashion items (\eg, ``is blue and has stripes''), but \ourdataset contains more generic and various instructions. Furthermore, manually annotated instructions can be inconsistent and noisy due to the inherent noisiness of crowdsourcing. For example, CIRR has instructions with no information (\eg, ``same environment different species''), or instructions which ignore the original information (\eg, ``the target photo is of a lighter brown dog walking in white gravel along a wire and wooden fence''). On the other hand, \ourdataset instructions only have modification information. Hence, a model trained on our CIR triplets can learn meaningful and generalizable representations, showing great ZS-CIR performances. Compared to the synthetic dataset of IP2P, our generation process is more scalable due to the keyword-based diverse caption generation process: Our caption triplets are synthesized based on keywords, \ourdataset covers more diverse keywords than IP2P (47k vs. 586k as shown in \cref{tab:keyword_stats}).
As a result, \ourdataset contains more massive triplets (1M vs. 18M), and CIR models trained on \ourdataset achieve better CIR performances even in the same dataset scale (1M). 

\subsection{Keyword-based diverse caption generation}
\label{subsec:gen_keyword_supp}
As the first approach to generating caption triplets, we collect captions from the existing caption datasets and modify the captions by replacing the object terms in the captions, \eg, $\langle$``a strawberry tart is ...'', ``covert strawberry to pak choi'', ``a pak choi tart is ...''$\rangle$ in \cref{fig:dataset_generation_overview}. For the caption dataset, 
We use the captions from COYO 700M \citep{kakaobrain2022coyo-700m}, StableDiffusion Prompts\footnote{\url{https://huggingface.co/datasets/Gustavosta/Stable-Diffusion-Prompts}} (user-generated prompts that make the quality of StableDiffusion better), LAION-2B-en-aesthetic (a subset of LAION-5B \citep{schuhmann2022laion}) and LAION-COCO datasets \citep{laioncoco} (synthetic captions for LAION-5B subsets with COCO style captions \citep{cococaption}. LAION-COCO less uses proper nouns than the real web texts).

We extract the object terms from the captions using the part-of-speech (POS) tagger provided by \texttt{Spacy}
\footnote{\url{https://spacy.io/}}. After frequency filtering, we have 586k unique object terms (\cref{tab:keyword_stats}). To make the caption triplet \rawcaptriplet, we replace the object term of each caption with other similar keywords by using the CLIP similarity score. More specifically, we extract the textual feature of keywords using the CLIP ViT-L/14 text encoder \citep{radford2021clip}, and we choose an alternative keyword from keywords with a CLIP similarity between 0.5 and 0.7. By converting the original object to a similar object, we have caption pairs of \rawcappair.

Using the caption pair \rawcappair, we generate the modification instruction text \rawcondtext based on a randomly chosen template from 48 pre-defined templates shown in \cref{tab:example_templates_supp}. After this process, we have the triplet of \rawcaptriplet. We generate $\approx$30M caption triplets by the keyword-based method.
\input{tables/keywords.tex}

\subsection{Amplifying InstructPix2Pix (IP2P) triplets by LLM}
\label{subsec:gen_llm_supp}
We also re-use the generated \rawcaptriplet by IP2P. We amplify the number of IP2P triplets by applying the efficient LoRA fine-tuning \citep{hu2021lora} to OPT-6.7B \citep{zhang2022opt} on the generated 452k caption triplets provided by \citet{brooks2022instructpix2pix}. Using the fine-tuned OPT, we generate $\approx$30M caption triplets.

\subsection{Triplet generation from caption triplets}
\label{subsec:triplet_gen_supp}

\begin{figure}[t]
    \centering
    \begin{minipage}[c]{0.48\textwidth}
    \centering
        \includegraphics[width=.75\textwidth]{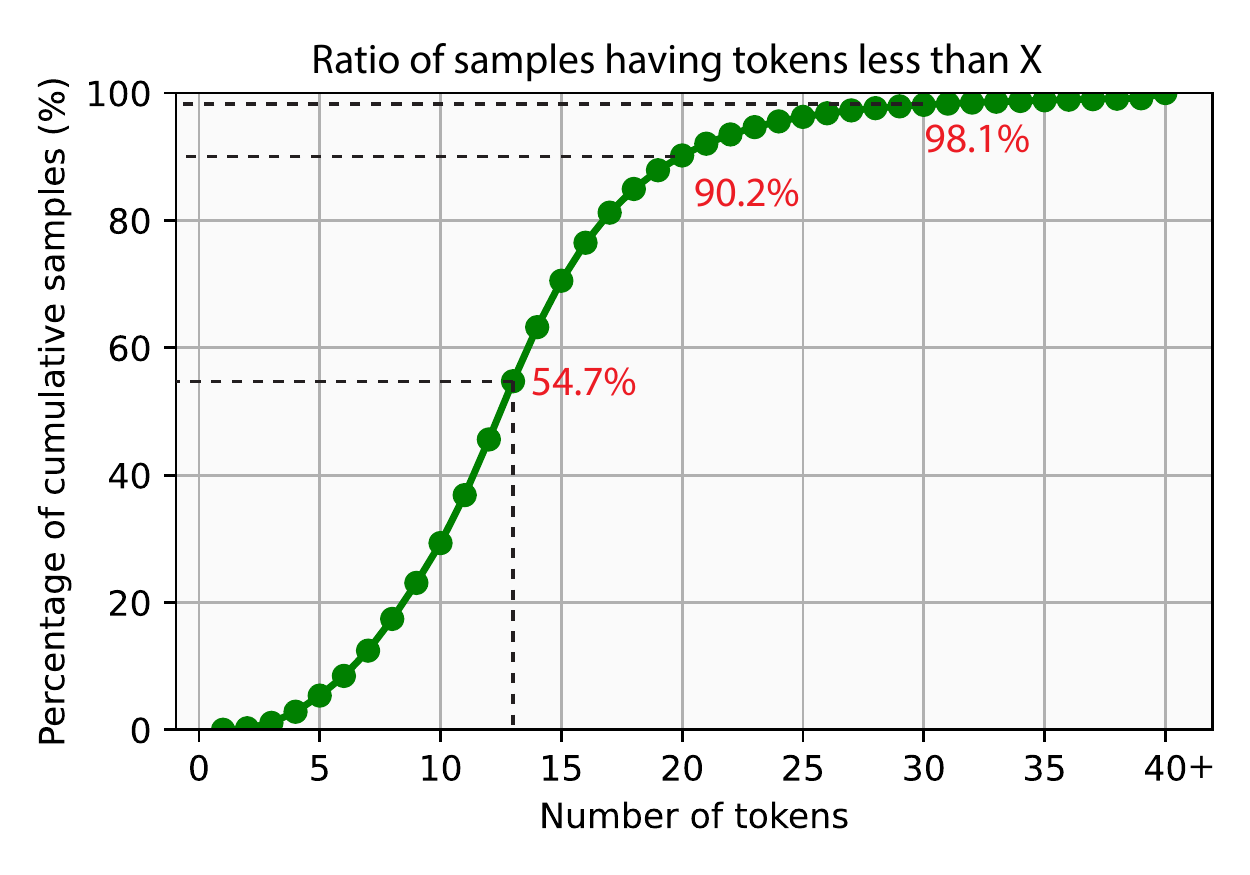}
        \vspace{-1.25em}
        \caption{\small {\bf Statistics of \ourdataset instructions.} We show the population of our instruction captions (\condtext) by the number of tokens per caption. We include captions having larger than 40 tokens in ``40+''.}
    \label{fig:data_stats_supp}
    \end{minipage}
    \hfill
    \begin{minipage}[c]{0.48\textwidth}
    \centering
        \includegraphics[width=.75\textwidth]{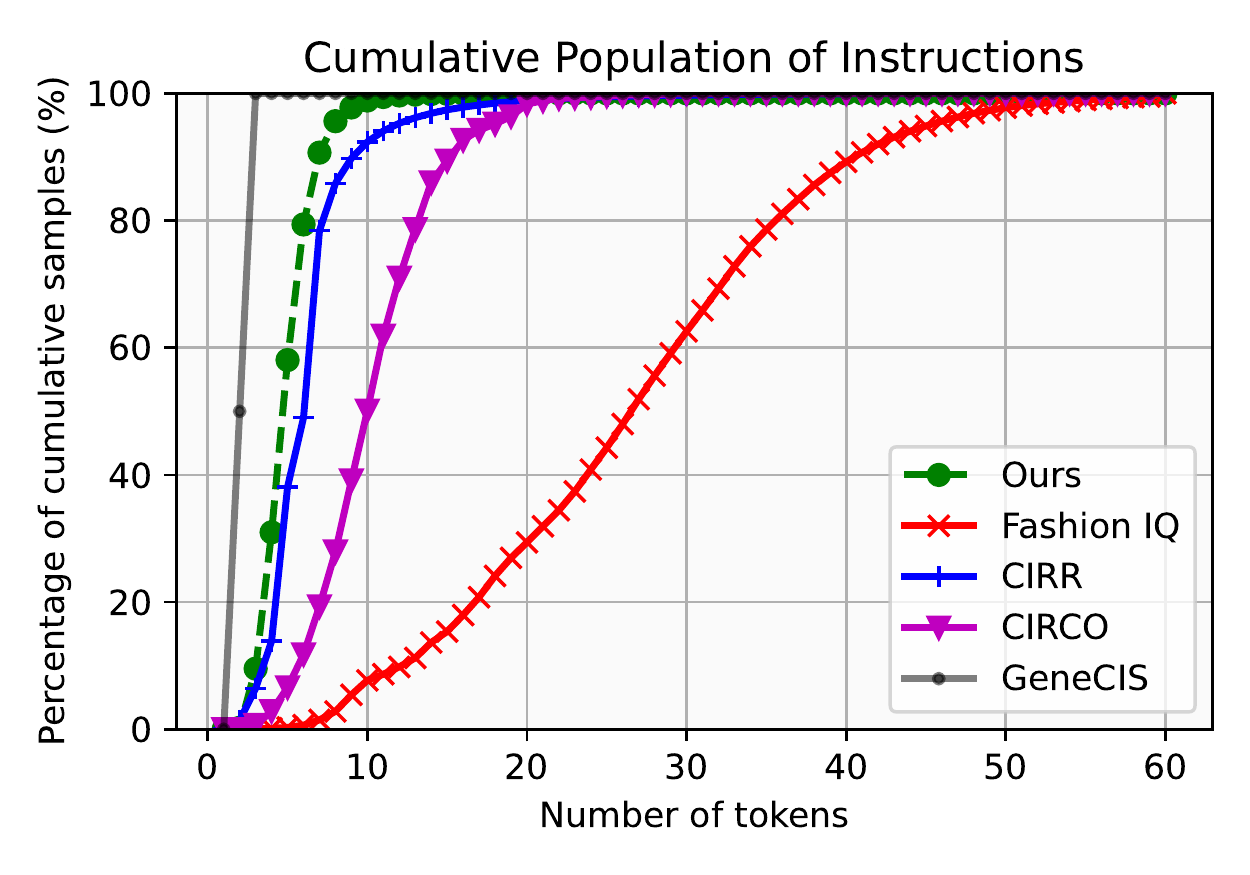}
        \vspace{-1.25em}
        \caption{\small {\bf Statistics of instructions of the CIR datasets.} We show the population of instruction captions (\eg, ``change A to B'') by the number of tokens. We include captions having larger than 60 tokens in ``60''.}
    \label{fig:instruction_stats_supp}
    \end{minipage}
    \vspace{-.5em}
\end{figure}

We generate 60M caption triplets \rawcaptriplet by the keyword-based generation process (\cref{subsec:gen_keyword_supp}) and the LLM-based generation process (\cref{subsec:gen_llm_supp})
(See \cref{subsec:dataset_stats_supp} for the statistics of the captions).
We generate images for \rawreftxt (original caption) and \rawtartxt (modified caption) using state-of-the-art text-to-image generation models, such as StableDiffusion (SD) 1.5, 2.0, 2.1, and SD Anime.
Following \citet{brooks2022instructpix2pix}, we apply Prompt-to-Prompt \citep{hertz2022prompttoprompt}, which aims to generate similar images while keeping the identity of the original image (\eg, the examples in \cref{fig:dataset_generation_overview}).
As a result, we generate 60M \rawtriplet (\condtext is given; \rawrefimg and \rawtarimg are generated by \rawreftxt and \rawtartxt, respectively).
While IP2P generates the samples only using SD 1.5, our generation process uses multiple DMs, for more diverse images not biased towards a specific model.

\subsection{CLIP-based filtering}
\label{subsec:filtering_supp}

Our generation process can include low-quality triplets, \eg, broken images or non-related image-text pairs. To prevent the issue, we apply a filtering process following \citet{brooks2022instructpix2pix} to remove the low-quality \rawtriplet. First, we filter the generated images for an image-to-image CLIP threshold of 0.70 (between \rawrefimg and \rawtarimg) to ensure that the images are not too different, an image-caption CLIP threshold of 0.2 to ensure that the images correspond to their captions (\ie, between \rawreftxt and \rawrefimg, and between \rawtartxt and \rawtarimg), and a directional CLIP similarity \citep{gal2022stylegan} of 0.2 ($L_\text{direction} := 1 - \text{sim}(\rawrefimg, \rawtarimg) \cdot \text{sim}(\rawreftxt, \rawtartxt)$, where $\text{sim}(\cdot)$ is the CLIP similarity) to ensure that the change in before/after captions correspond with the change in before/after images. For keyword-based data generation, we filter out for a keyword-image CLIP threshold of 0.20 to ensure that images contain the keyword (\eg, image-text CLIP similarity between the strawberry tart image and the keyword ``strawberry'' in \cref{fig:dataset_generation_overview}).
For instruction-based data generation, we filter out for an instruction-modified image CLIP threshold of 0.20 to ensure consistency with the given instructions.

The filtering process prevents low-quality triplets, such as broken images. For example, if a modified caption does not make sense to generate a corresponding image (\eg, changing ``apples are on the tree'' to ``apples are on the thunderstorm''), then StableDiffusion and Prompt2Prompt will not be able to generate proper images for the modified caption. These images will be filtered out by CLIP similarity because we measure the similarity between the generated images is sufficiently high, and the broken images will have low similarities with clean images. Similarly, if the original caption is not suitable for generating a corresponding image (\eg, ``invisible strawberry''), then it will fail to pass the keyword-image CLIP filtering.

\begin{figure}[h!]
    \centering
    \includegraphics[width=\linewidth]{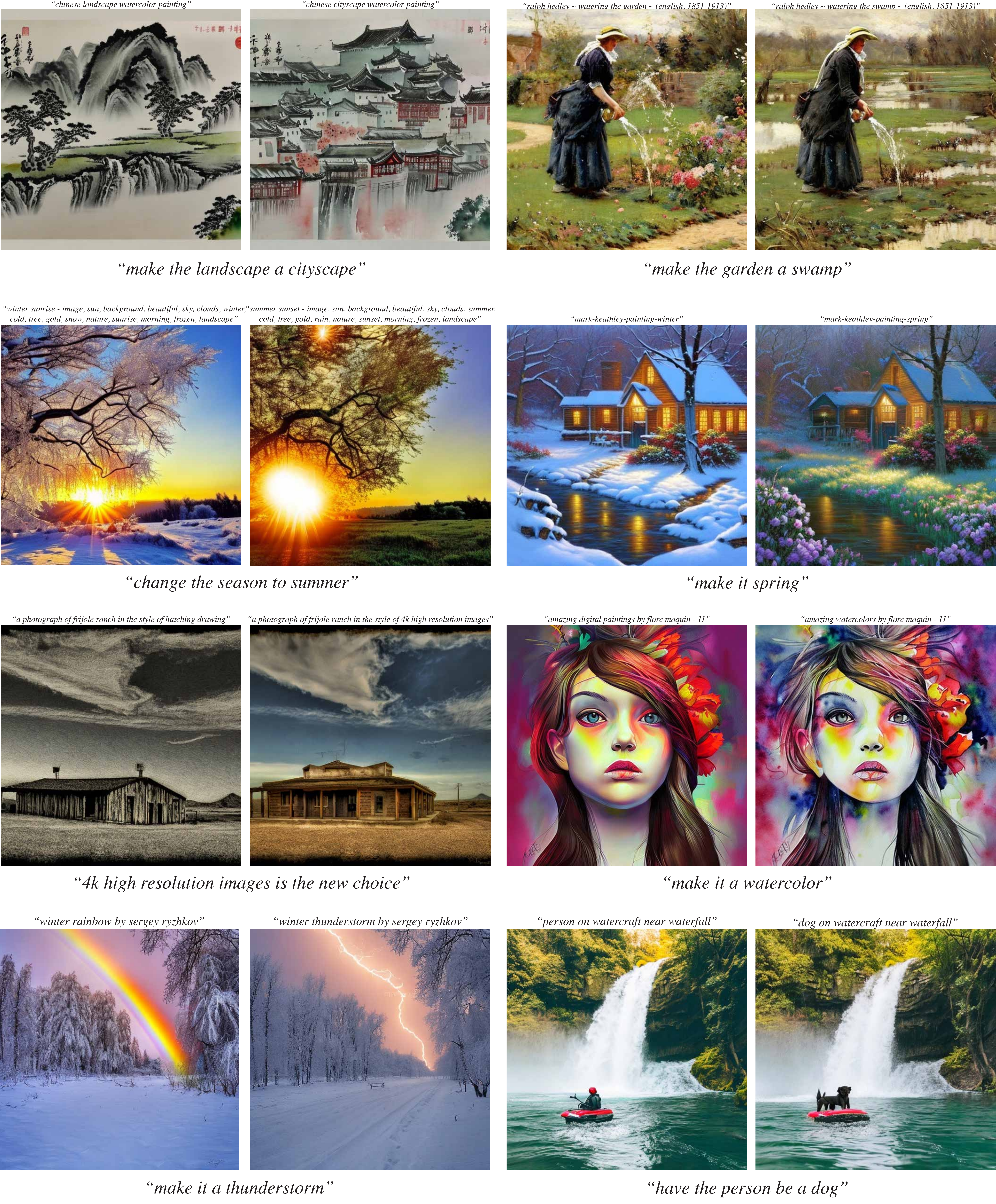}
    \caption{\small {\bf Examples of \ourdataset.} We show examples of \rawtriplet, \ie, \{original image, modification instruction, and modified image\}, as well as the generation prompt for \rawrefimg and \rawtarimg.}
    \label{fig:dataset_examples}
    \vspace{-1em}
\end{figure}

After the filtering, we have 11.4M \rawtriplet from the keyword-based generated captions and 7.4M \rawtriplet from the LLM-based generated captions. It implies that the fidelity of our keyword-based method is higher than OPT fine-tuning in terms of T2I generation. As a result, \ourdataset contains 18.8M synthetic \rawtriplet.
Examples of our dataset are shown in \cref{fig:dataset_examples}.

\subsection{Dataset Statistics}
\label{subsec:dataset_stats_supp}

We show the statistics of our generated caption dataset (\ie, before T2I generation, \rawreftxt and \rawtartxt). We use the CLIP tokenizer to measure the statistics of the captions. \cref{fig:data_stats_supp} shows the cumulative ratio of captions with tokens less than X. About half of the captions have less than 13 tokens, and 90\% of the captions have less than 20 tokens. Only 0.8\% of the captions have more than 40 tokens.

We also compare \ourdataset, FashionIQ, CIRR, CIRCO, and GeneCIS in the token statistics of instructions (\ie, \rawcond). 
\cref{fig:instruction_stats_supp} shows that the instruction statistics vary across different datasets. We presume that this is why the ZS-CIR still has difficulty outperforming the task-specific supervised CIR methods.

\ifx\preprint\undefined
\subsection{Example samples of \ourdataset}
\label{subsec:dataset_examples}

We illustrate example samples of \ourdataset in \cref{fig:dataset_examples}. Our dataset can express the change of overall context (\eg, ``make the landscape a cityscape''), the seasonal change (\eg, ``make it sprint''), the change of mood (\eg, ``make it a watercolor''), and the change of local objects (\eg, ``have the person be a dog'').
\textbf{The full dataset will be hosted through the HuggingFace hub}, and samples are in the supplementary materials.
\else
\fi

\section{Experiments}

\subsection{Implementation details}

\textbf{Encoders.}
We use three different CLIP models for image encoder (\cref{fig:method_inference_overview} ``CLIP Img Enc''), the official CLIP ResNet-50 and ViT-L/14 \citep{radford2021clip}, and CLIP ViT-G/14 by OpenCLIP \citep{openclip}, whose feature dimensions are 768, 1024, and 1280, respectively.
Beyond the backbone size, we observe the choice of the text condition encoder is also important (\cref{fig:method_inference_overview} ``Txt Enc'').
As shown \citet{balaji2022ediffi}, using a text-oriented model such as T5 \citep{raffel2020t5} in addition to the CLIP textual encoder results in improved performance of text-to-image generation models. Motivated by this observation, we also use both the CLIP textual encoder and the language-oriented encoder for small image encoder (\ie, CLIP ViT-L/14). We also observed the positive effect of the text-oriented model and experiment results showed that T5-XL, which has 3B parameters, could improve the performance by a large margin in the overall evaluation metrics.

\textbf{Denoiser.}
We use a simple Transformer architecture for the denoising procedure, instead of the denoising U-Net \citep{rombach2022latentdiffusion}. We empirically observe that our transformer architecture performs slightly better than the U-Net architecture, but is much simpler. We use the multi-head self-attention blocks as the original Transformer \citep{vaswani2017attention}. We set the depth, the number of heads, and the dimensionality of each head to 12, 16, and 64, respectively. The hidden dimension of the Transformer is set to 768 and 1280 for ViT-L and ViT-G, respectively.
The denoising Transformer takes two inputs: a noisy visual embedding and a time-step embedding. The conditions (\eg, text, mask and image conditions) are applied only to the cross-attention layer; thereby it is computationally efficient even using many conditions. \ours is similar to the ``DiT with cross-attention'' by \citet{peebles2022scalable}, but handles more various conditions.

\textbf{Training details.}
For the efficient training, all visual features are pre-extracted and frozen.
All training text embeddings are extracted at every iteration. To improve computational efficiency, we reduced the number of input tokens of the T5 models to 77, as in CLIP.
A single-layer perceptron was employed to align the dimension of text embeddings extracted from T5-XL with that of CLIP ViT-L/14.

\subsection{Experiment settings}

All models were trained using AdamW \citep{loshchilov2017decoupled}. We used DDIM \citep{song2020denoising} for the sampling variance method. We did not apply any image augmentation but used pre-extracted CLIP image features for computational efficiency; text features were extracted on the fly as text conditions can vary in \ourdataset.
We report the detailed hyperparameters in \cref{tab:parameters}.

We evaluate the zero-shot (ZS) capability of \ours on four CIR benchmarks, including FashionIQ \citep{fashioniq}, CIRR \citep{cirr}, CIRCO \citep{circo} and GeneCIS \citep{genecis}. 
We compare \ours to the recent ZS CIR methods, including Pic2Word \citep{saito2023pic2word} and SEARLE \citep{circo}. We also reproduce the fusion-based methods, such as ARTEMIS \citep{delmas2022artemis} and Combiner \citep{baldrati2022clip4cir}, on \ourdataset and report their ZS performances.
Note that the current CIR benchmarks are somewhat insufficient to evaluate the effectiveness of \ours, particularly considering real-world CIR queries. More details are found in \cref{subsec:dataset_details_supp}.
Our work is the first study that shows the impact of the dataset scale and the zero-shot CIR performances with various methods, such as our method, ARTEMIS and Combiner.
Due to the page limit, the details of each task and method are in \cref{subsec:baselines_supp} and \ref{subsec:dataset_details_supp}.
We also compare our training strategy with the other CIR methods in terms of the complexity in \cref{subsec:baselines_supp}. In summary, we argue that our method is not specifically complex compared to the other methods.
The fine-tuned performances on FashionIQ and CIRR are in \cref{subsec:full_experiment_results_supp}.

\subsection{Comparison methods.}
\label{subsec:baselines_supp}
We compare \ours with four state-of-the-art CIR methods as follows:

\textbf{Combiner} \citep{baldrati2022clip4cir} involves a two-stage training process. First, the CLIP text encoder is fine-tuned by contrastive learning of \cond and \refimg $+$ \cond in the CLIP embedding space. The second stage replaces \refimg $+$ \cond to the learnable Combiner module. Only the Combiner module is trained during the second stage.

\textbf{ARTEMIS} \citep{delmas2022artemis} optimizes two similarities simultaneously. The implicit similarity is computed between the combined feature of \tarimg and \cond, and the combined one of \refimg and \cond. The explicit matching is computed between \tarimg and \cond. ARTEMIS suffers from the same drawback as previous CIR methods, \eg TIRG \citep{vo2019tirg}: As it should compute combined feature of \tarimg and \cond, it is not feasible to use an approximate nearest neighbor search algorithm, such as FAISS \citep{faiss}. This is not a big problem in a small dataset like FashionIQ, but it makes ARTEMIS infeasible in real-world scenarios, \eg, when the entire LAION-2B is the target database.
As Combiner and ARTEMIS are trained based on CLIP ResNet-50 (RN50), we also train \ours based on the same RN50 features for a fair comparison.

\textbf{Pic2Word} \citep{saito2023pic2word} projects a visual feature into text embedding space, instead of combining them. Pic2Word performs a text-to-image retrieval by using the concatenated feature as the input of the CLIP textual encoder. As the projection module is solely trained on cheap paired datasets without expensive triplet datasets, it is able to solve CIR in a zero-shot manner.

\textbf{SEARLE} \citep{circo} is a similar to Pic2Word, while SEARLE employes text-inversion task instead of projection. \citet{circo} also proposed a technique, named Optimization-based Textual Inversion (OTI), a GPT-powered regularization method. In this paper, we compare SEARLE-XL and SEARLE-XL-OTI, which use the ViT-L/14 CLIP backbone for a fair comparison.

We also compare a naive editing-based retrieval method: editing an image using IP2P with an instruction and retrieving images using the CLIP image encoder.
Note that Pic2Word and SEARLE are trained on the image-only datasets; therefore, it is impossible to train them on \ourdataset. Similarly, Combiner and ARTEMIS only take triplets for training; hence, they cannot be trained with image-caption datasets, such as Conceptual Captions \citep{sharma2018conceptual} or LAION \citep{schuhmann2022laion}. Due to this reason, we only train the previous fusion-based methods, ARTEMIS and Combiner on \ourdataset for comparison. Note that Combiner needs a pre-trained CLIP model trained on a large-scale image-caption dataset. In this point of view, we can argue that Combiner and CompoDiff are comparable in terms of the dataset scale; both methods use a large-scale image-caption dataset and synthetic triplets in our experiment.

\subsection{CIR Datasets.}
\label{subsec:dataset_details_supp}

\textbf{FashionIQ} \citep{fashioniq} has (46.6k / 15.5k / 15.5k) (training / validation / test) images with three fashion categories: Shirt, Dress, and Toptee.
Each category has 18k training triplets and 12k evaluation triplets of \rawtriplet. Examples of $x_c$ looks like: ``is more strappy and emerald'', ``is brighter'' or ``is the same''. The main drawback of FashionIQ is that the dataset is limited to the specific subsets of fashion domains, hence it is not possible to evaluate whether the methods are truly useful for real-world CIR tasks.

\textbf{CIRR} \citep{cirr} contains more generic images than FashionIQ. CIRR uses the images from NLVR$^2$ \citep{suhr2018corpus} with more complex and long descriptions. CIRR has 36k open-domain triplets divided into the train, validation, and test sets in 8:1:1 split.
The examples of $x_c$ are ``remove all but one dog and add a woman hugging it'', ``It's a full pizza unlike the other with the addition of spice bottles'', or ``Paint the counter behind the brown table white''.
As the example captions show, one of the main drawbacks of CIRR is that the text queries are not realistic compared to the real user scenario; we may need more shorter and direct instructions. As the text instruction distribution of CIRR is distinct from others (\cref{fig:instruction_stats_supp}), we observe that sometimes CIRR results are a not reliable measure of open-world CIR tasks (\eg, retrieval on LAION).
\citet{circo} also observe another main drawback of CIRR. CIRR contains a lot of false negatives (FNs) due to the nature of the image-text dataset \citep{chun2021pcme, chun2022eccv_caption, chun2023pcmepp}

\textbf{CIRCO.}
To tackle the FN issue of CIRR, \citet{circo} introduces the CIRCO dataset. This dataset comprises of 1020 queries, where 220 and 800 of them are used for validation and test, respectively. The images are from the COCO dataset \citep{lin2014coco}, where the size of the image database is 120K COCO images.
Example $x_c$ of CIRCO includes ``has two children instead of cats'' or ''is on a track and has the front wheel in the air``. We use mAP scores following \citet{circo} which is known to be a robust retrieval metric \citep{musgrave2020metric,chun2022eccv_caption}.

\textbf{GeneCIS} \citep{genecis} is a dataset to evaluate different conditional similarities of four categories: (1) focus on an attribute, (2) change an attribute, (3) focus on an object, and (4) change an object.
The first two categories are based on the VAW dataset \citep{vaw}, which contains massive visual attributes in the wild. The other categories are sampled from COCO.

\textbf{The limitations of the current CIR benchmarks.}
One of the significant drawbacks of the existing CIR benchmarks, such as FashionIQ \citep{fashioniq} and CIRR \citep{cirr}, is the domain-specific characteristics that cannot be solved without training on the datasets. For example, the real-world queries that we examine are mainly focused on small editing, addition, deletion, or replacement. However, because the datasets are constructed by letting human annotators write a modification caption for a given two images, the text conditions are somewhat different from the real-world CIR queries. For example, CIRR dev set has some text conditions like: ``show three bottles of soft drink'' (different from the common real-world CIR text conditions), ``same environment different species'' (ambiguous condition), ``Change the type of dog and have it walking to the left in dirt with a leash.'' (multiple conditions at the same time). These types of text conditions are extremely difficult to solve in a zero-shot manner, but we need access to the CIRR training dataset.
Instead, we believe that CIRCO is a better benchmark than FashionIQ and CIRR. It is because CIRCO aims to resolve the false negatives of FashionIQ and CIRR, focusing on the retrieval evaluation.

Furthermore, the existing benchmarks (even more recent ones, such as CIRCO and GeneCIS) cannot evaluate the negative and mask conditions.
In practice, when we perform a qualitative study on a large image index (\ie, LAION-2B image index), we observe that \ours outperforms previous methods, such as Pic2Word, in terms of the retrieval quality (See \cref{fig:ours_vs_pic2word} and \cref{fig:ours_vs_pic2word_more_supp} for examples).
Also, we observe that even if the CIR scores are similar (\eg, 10M and 18.8M in \cref{tab:dataset_scale}), in the LAION-2B index, the retrieval quality can be significantly better.
Unfortunately, it is impossible to evaluate the quality of retrieval results on the LAION-2B, because a quantitative study requires expensive and infeasible human verification.

\input{tables/main_table}

\subsection{Qualitative comparisons on four Zero-shot CIR (ZS-CIR) benchmarks}

\cref{tab:main} shows the overview of ZS-CIR comparison results. CLIP + IP2P denotes the naive editing-based approach by editing the reference image with the text condition using IP2P and performing image-to-image retrieval using CLIP ViT-L. In the table, \ours outperforms all the existing methods with significant gaps. The table shows the effectiveness both of our diffusion-based CIR approach and our massive synthetic dataset. In the \ourdataset-trained group, \ours outperforms previous SOTA fusion-based CIR methods with a large gap, especially on CIRR and CIRCO, which focus on real-life images and complex descriptions. Our improvement is not main due to the architecture, as \ours already outperforms the fusion methods in RN50. We also can observe that the \ourdataset-trained group also enables the fusion-based methods to have the zero-shot capability competitive to the SOTA zero-shot CIR methods, Pic2Word and SEARLE. More interestingly, ARTEMIS even outperforms its supervised counterpart on FashionIQ with a large gap (40.6 vs. 38.2 in average recall).
We provide the full results of \cref{tab:main}, the supervised results of ARTEMIS, Combiner, and \ours trained on \ourdataset in \cref{subsec:full_experiment_results_supp}.

Compared to the previous ZS-CIR methods (Pic2Word and SEARLE), \ours achieves remarkable improvements on the same architecture scale (\ie, ViT-L), except on CIRR. We argue that it is due to the noiseness of the CIRR dataset. Instead, \ours outperforms the other methods on FashionIQ, CIRCO and GeneCIS with a significant gap. We believe that it is because \ours explicitly utilizes the diverse and massive synthetic triplets, while Pic2Word and SEARLE only employ images and the ``a photo of'' caption during training, resulting in a lack of diversity and generalizability.

\subsection{Inference time}
\label{subsec:inference_time_supp}

One possible drawback of \ours is a relatively slow inference compared to encoder-only methods. However, we confirm that \ours is practically useful with high throughput (120 $\sim$ 230ms) for a single image (\cref{tab:denoising_steps}). The table reports throuputs of the comparison methods on ViT-L/14 backbone (Pic2Word, SEARLE, \ours) or RN50 backbone (ARTEMIS, Combiner).
One of the advantages of \ours is that we can control the trade-off between retrieval performance and inference time. Note that it is impossible for the other methods. If we need a faster inference time, even with a worse retrieval performance, we can reduce the number of diffusion steps as shown in \cref{tab:denoising_steps}.
Even with only 4 or 5 iterations, our model can produce competitive results. If we use 100 steps, we have a slightly better performance (42.65 vs. 42.17), but a much slower inference time (2.02 sec vs. 0.12 sec). In the experiments, we set the step size to 10.

\input{tables/ablation_results_per_steps}

Our another contribution is a novel and efficient conditioning for the diffusion model. Instead of using a concatenated vector of all conditions and inputs as the input of the diffusion model, we use the cross-attention mechanism for conditions and leave the input size the same as the original size. As shown in the next section (\cref{tab:i2t_retrieval}), our design choice is three times faster than the naive implementation.

We also remark that recently, there have been huge advances in boosting diffusion model inference speed, such as the Consistency Model \citep{song2023consistency} or distillation \citep{meng2023distillation}. Using more efficient diffusion model variants for the CompoDiff model will also greatly improve the inference speed. However, these approaches will need a completely different training framework \citep{song2023consistency} or a complicated two-staged distillation process \citep{meng2023distillation}. While we believe that applying these techniques to CompoDiff will bring a huge benefit, we leave further improvements for future work.

\subsection{Analysis}
\label{subsec:abl}

\paragraph{Denoising Transformer ablation.}
\input{tables/i2t_retrieval}
\ours does not take textual embeddings \condtext as concatenated input tokens of the denoising Transformer but as a condition of cross-attention for the efficient inference. We compare the impact of different design choices for handling textual embeddings. First, we evaluate the Dall-e2 ``Prior'' model \citep{dall-e2} which converts CLIP textual embeddings into CLIP visual embeddings (we use a public community model \citep{kakaobrain2022karlo-v1-alpha} because the official model is not yet publically available). Second, we test the ``Prior-like'' model by using the denoising Transformer, but taking text guidance as input tokens instead of cross-attention. We also test two more \ours models from our two-stage training strategy.

As the Prior models are not designed for handling CIR triplets, we measure their ability on image-to-text (I2T) and text-to-image (T2I) cross-modal retrieval benchmarks on the MS-COCO Caption dataset \citep{cococaption}. We also evaluate them on the extension of COCO Caption to mitigate the false negative problem of COCO, CxC \citep{cxc} and ECCV Caption \citep{chun2022eccv_caption}. \cref{tab:i2t_retrieval} shows the average I2T and T2I metrics of each benchmark. In the table, we first observe that our design choice is three times faster than the ``Prior-ish'' counterparts by handling textual embeddings with cross-attention. Second, we observe that Stage 1 only \ours shows a better understanding of I2T and T2I tasks. We speculate that this is because Ours (Stage 1 only) is directly optimized by the image-to-text (ITM) matching style dataset, while Ours (Stage 1 + Stage 2) is also trained with other types of conditions (\eg, masks, negative texts, image conditions).
In summary, our design choice shows $\times$ 3 faster inference time than the prior model \citep{dall-e2} but better cross-modal retrieval performances on COCO and its extensions.

\input{tables/ablation_dataset_scale}
\paragraph{Impact of dataset scale.}
\cref{tab:dataset_scale} shows the impact of the dataset scale by \ourdataset on ARTEMIS, Combiner and \ours.
First, at a scale of 1M, models trained on our 1M subset significantly outperformed the IP2P triplets. This result indicates that our dataset has a more diverse representation capability. As the size of our dataset increases, the performance gradually improves. 
Notably, \ourdataset shows consistent performance improvements from 1M to 18.8M, where manually collecting triplets in this scale is infeasible and nontrivial. Thanks to our diversification strategy, particularly keyword-based generation, we can scale up the triplet to 18.8M without manual human labor.

\cref{tab:dataset_scale} shows that the massive data points are not the necessary condition for training \ours,
but all methods are consistently improved by scaling up the data points.
It is also worth noting that although the FashionIQ and CIRR scores look somewhat saturated after 10M, these scores cannot represent authentic CIR performances due to the limitations of the datasets as discussed in \cref{subsec:dataset_details_supp}.
As far as we know, this is the first study shows the impact of the dataset scale to the zero-shot CIR performances.

\begin{wraptable}{r}{0.4\linewidth}
\small
\centering
\setlength{\tabcolsep}{4pt}
\begin{tabular}{l|ccc}
\toprule
Text Enc & T5-XL & CLIP & CLIP + T5-XL \\
\midrule
FashionIQ & 38.20 & 42.33 & \textbf{44.11} \\
CIRR & 29.19 & 37.83 & \textbf{39.25} \\
\bottomrule
\end{tabular}
\caption{\small \textbf{Impact of text encoder.}}
\label{tab:improved_text_encoder}
\vspace{-1em}
\end{wraptable}
\paragraph{Condition text encoder.}
As observed by \citet{balaji2022ediffi} and \citet{byun2024rtd}, the power of the CLIP textual encoder affects a lot to the performance of image-text tasks. Motivated by this observation, we also use both the CLIP textual encoder and the language-oriented encoder for extracting the text features of \cref{eq:stage2-triplet}. \cref{tab:improved_text_encoder} shows the choice of the text encoder affects a lot to the performance on ViT-L/14 CLIP backbone. When the CLIP textual encoder and the T5-XL were used together, the results improved significantly. We suspect that this is because the strong T5 encoder can help the CLIP text encoder to better understand given captions. Interestingly, we observe that using T5 alone degrades the performance even compared to using the CLIP textual encoder alone. We presume this is because T5-XL is specified for long text sequences (\eg, larger than 100 tokens) and text-only data. Meanwhile, \ourdataset has a short average length (See \cref{fig:data_stats_supp} and \ref{fig:instruction_stats_supp}), which is not specialized by T5. Also, our dataset is based on captions, paired with an image; we also need to consider image information to understand the given caption, but we cannot handle image information alone with T5.

Therefore, we use both T5-XL encoder \citep{raffel2020t5} and CLIP text encoder for the ViT-L model. For the ViT-G model, we use the CLIP text encoder only because we empirically observe that the textual representation power of the ViT-G is much more powerful than the ViT-L.
Exploring the power of textual representations, such as learnable prompts to the CFG guidance, could be an interesting future work.

\input{tables/ablation_stage2}

\paragraph{Stage 2 ablation.}
As we described in \cref{subsec:training}, we alternatively update the model using three different objectives in Stage 2. Here, we conduct an ablation study of our design choice in \cref{tab:ablation_stage2}. Our alternative learning strategy improves the overall performance. It is because although \ourdataset is a vast and diverse dataset, its diversity is weaker than LAION. While we train \ours to handle triplets by training on \ourdataset using \cref{eq:stage2-triplet}, we employ additional tasks using LAION, \ie, \cref{eq:stage1-t2i} and \cref{eq:stage2-inpaint} for achieving a better generalizability.

\begin{wrapfigure}{r}{0.3\textwidth}
\centering
\vspace{-3.75em}
\includegraphics[width=.85\linewidth]{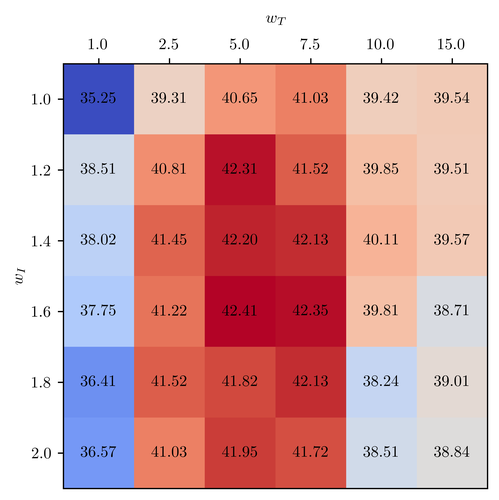}
\vspace{-.75em}
\caption{\small {\bf $w_I$ and $w_T$ vs. FashionIQ ZS-CIR performance.}}
\label{fig:heatmaps}
\vspace{-2.5em}
\end{wrapfigure}
\paragraph{The choice of $w_I$ and $w_T$.}
We include the retrieval performances by varying conditions in \cref{fig:heatmaps}. As $w_I$ increases, the generated image embeddings become more dependent on the reference image, while increasing $w_T$ results in a greater influence of the text guidance (See \cref{fig:cfg_variation_images}). However, large $w_I$ and $w_T$ are not always beneficial. If $w_I$ or $w_T$ is too large, it can lead to unexpected results. To find a harmonious combination of $w_I$ and $w_T$, we performed a sweeping process as shown in \cref{fig:heatmaps}. We use $w_I$ as 1.5 and $w_T$ as 7.5 considering the best content-condition trade-off.

\begin{figure*}[t]
    \centering
    \includegraphics[width=.9\linewidth]{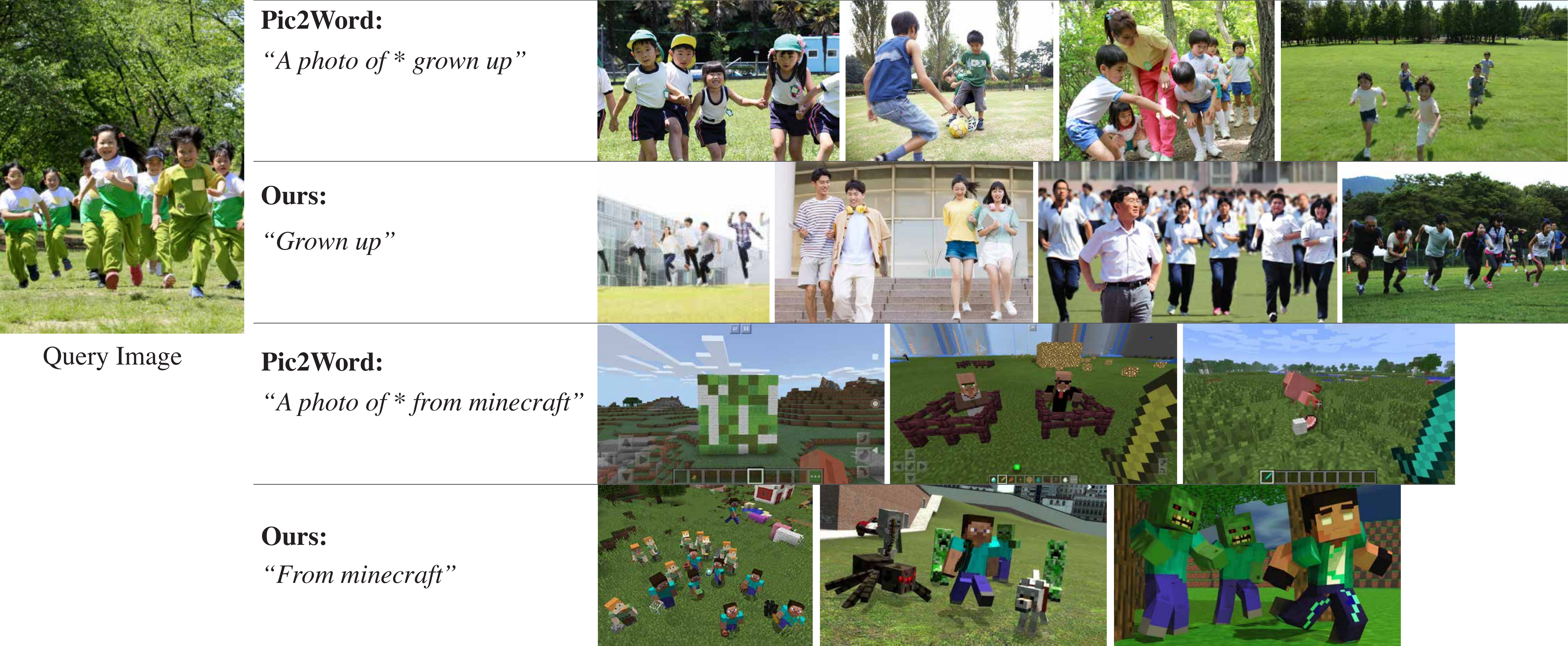}
    \vspace{-.5em}
    \caption{\small {\bf Qualitative comparison of zero-shot CIR for Pic2Word and \ours.} We conduct CIR on LAION. As Pic2Word cannot take a simple instruction, we made a simple modification for the given instruction.}
    \label{fig:ours_vs_pic2word}
\vspace{-.5em}
\end{figure*}
\subsection{Qualitative examples}
\label{subsec:examples}
We qualitatively show the versatility of \ours for handling various conditions. For example, \ours not only can handle a text condition, but it can also handle a \textit{negative} text condition (\eg, removing specific objects or patterns in the retrieval results), masked text condition (\eg, specifying the area for applying the text condition). \ours even can handle all conditions simultaneously (\eg, handling positive and negative text conditions with a partly masked reference image at the same time). To show the quality of the retrieval results, we conduct a zero-shot CIR on entire LAION dataset \citep{schuhmann2022laion} using FAISS \citep{faiss} for simulating billion-scale CIR scenarios.

\cref{fig:ours_vs_pic2word} shows qualitative comparsions of zero-shot CIR results by Pic2Word and \ours. \ours results in semantically high-quality retrieval results (\eg, understanding the ``crowdedness'' of the query image and the meaning of the query text at the same time). However, Pic2Word shows poor understanding of the given queries, resulting in unfortunate retrieval results (\eg, ignoring ``grown up'' of text query, or the  ``crowdedness'' of the query image). More examples are in \cref{fig:ours_vs_pic2word_more_supp}.

\begin{figure}[t]
    \centering
    \includegraphics[width=\linewidth]{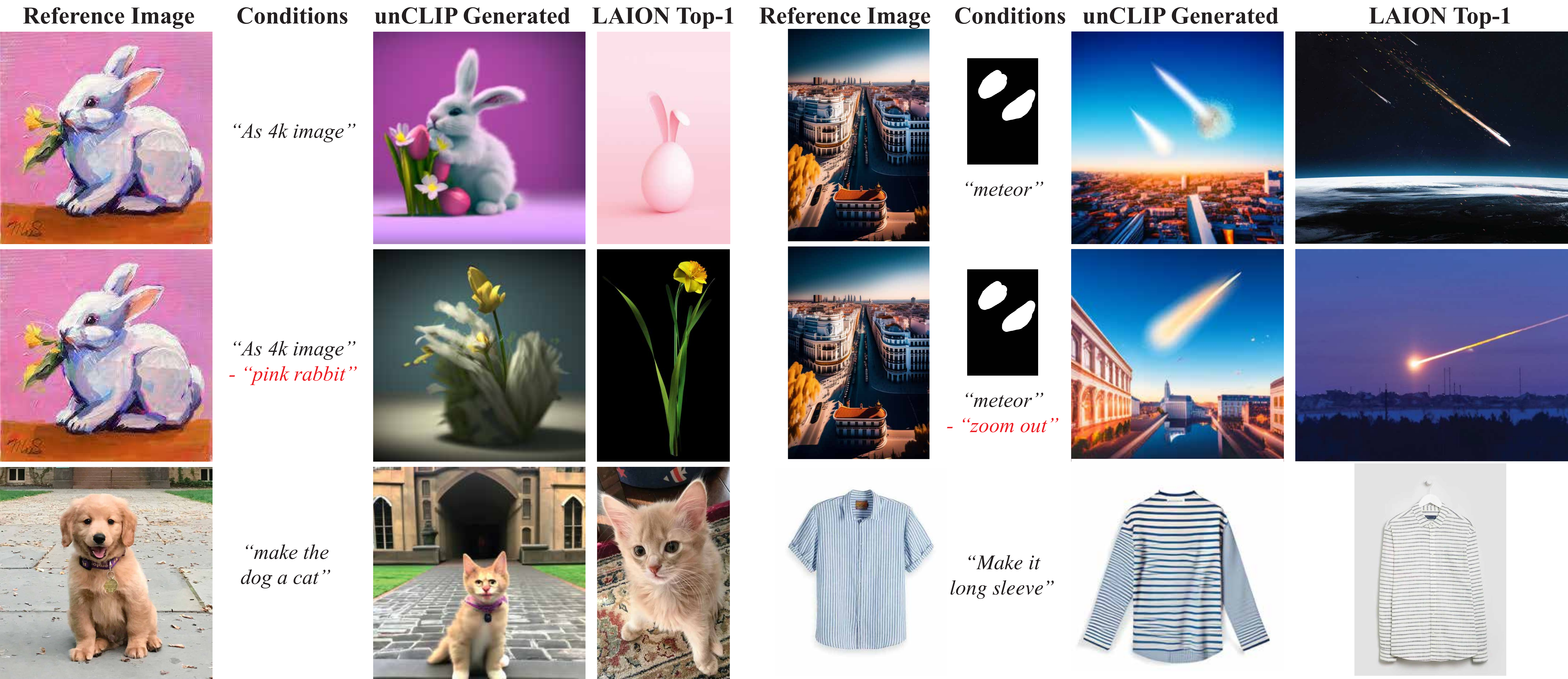}
    \caption{\small {\bf Generated and retrieved images by \ours.} Images are generated by unCLIP decoder and retrieved from LAION using transformed features by \ours. More examples are shown in \cref{subsec:unclip_supp}.}
    \label{fig:unclip_examples}
\vspace{-.5em}
\end{figure}

Finally, it is worth noting that \ours generates a feature belonging to the CLIP visual latent space.
Namely, unCLIP \citep{dall-e2}, which decodes a CLIP image feature to an image, can be applied to our composed features. We compare the top-1 retrieval results from LAION and the generated images in \cref{fig:unclip_examples} and \cref{subsec:unclip_supp}. We use the Karlo unCLIP decoder \citep{kakaobrain2022karlo-v1-alpha}, by replacing the original Prior module to \ours.
As shown in the figures, \ours can manipulate the given input reflecting the given conditions.
We believe that incorporating unCLIP into the real-world search engine could potentially improve the user experience by generating images when the desired search results are not available.

\section{Conclusion}
We have introduced \ours, a novel diffusion-based method for solving complex CIR tasks. We have created a large and diverse dataset named \ourdataset, consisting of 18.8M triplets of images, modification texts, and modified images. \ours has demonstrated impressive ZS-CIR capabilities, as well as remarkable versatility in handling diverse conditions, such as negative text or image masks, and the controllability to enhance user experience, such as adjusting image text query weights. Furthermore, by training the existing CIR methods on \ourdataset, the models became comparable ZS predictors to the ZS-CIR methods. We strongly encourage future researchers to leverage our dataset to advance the field of CIR.

\section*{Societal Impact}

Our work is primarily focused on solving complex composed image retrieval (CIR) challenges and is not designed for image editing purposes. However, we are aware that with the use of additional public resources, such as the community version of the unCLIP feature decoder \citep{dall-e2}, our method can potentially be utilized as an image editing method. We would like to emphasize that this unintended application is not the primary objective of our research, and we cannot guarantee the effectiveness or safety of our method in this context.
It is important to note that we have taken steps to mitigate potential risks associated with the unintended use of our method for image editing. For instance, we applied NSFW filters to filter out potentially malicious samples during the creation of \ourdataset. Nevertheless, we recognize the need for continued research into the ethical and societal implications of AI technologies and pledge to remain vigilant about potential unintended consequences of our work.

{\small
\bibliography{main}
\bibliographystyle{tmlr}
}

\clearpage
\appendix
\numberwithin{equation}{section}
\numberwithin{figure}{section}
\numberwithin{table}{section}

\section*{Appendix}

In this additional document, we provide the detailed training hyperparameter settings (\cref{subsec:parameters_supp}), the full experimental results (\cref{subsec:full_experiment_results_supp}) and more qualitative examples (\cref{subsec:more_examples_supp}).

\section{Hyperparameter details}
\label{subsec:parameters_supp}

\cref{tab:parameters} shows the detailed training hyperparameters of \ours.

\input{tables/parameters}

\section{Full experiment results of \cref{tab:main}}
\label{subsec:full_experiment_results_supp}

In this subsection, we report the full experiment results of \cref{tab:main}. For FashionIQ and CIRR datasets, we also report additional fine-tuning results of \ourdataset-trained CIR methods (\ie, directly fine-tune the ``zero-shot'' methods on the target training dataset).

\input{tables/fashion_iq_main}

\input{tables/cirr_main}

\textbf{FashionIQ.}
\cref{tab:fashioniq} shows the comparison of \ours with baselines on the FashionIQ dataset.
Following the standard choice, we use recall@K as the evaluation metric. ``Zero-shot'' means that the models are not trained on FashionIQ (\ie, same as \cref{tab:main}). ARTEMIS and Combiner were originally designed for the supervised setting, but, we trained them on \ourdataset for a fair comparison with our method. Namely, we solely train them on \ourdataset for the zero-shot benchmark and fine-tune the zero-shot weights on the FashionIQ training set for the supervised benchmark.

\textbf{CIRR.}
\cref{tab:cirr} shows the CIRR results and all experimental settings were identical to FashionIQ. Similar to FashionIQ, \ours also achieves a new state-of-the-art CIRR zero-shot performance.
It is noteworthy that Combiner performs great in the supervised setting but performs worse than \ours in the zero-shot setting. We presume that the fine-tuned Combiner text encoder is overfitted to long-tailed CIRR captions. It is partially supported by our additional experiments on text encoder in \cref{subsec:abl}; a better understanding of complex texts provides better performances.

For both datasets, we achieve the best scores by fine-tuning the Combiner model trained on \ourdataset to the target dataset. It shows the benefits of our dataset compared to the limited CIR triplet datasets.

\input{tables/circo_main}
\input{tables/genecis_main}

\textbf{CIRCO.}
The detailed CIRCO results are shown in \cref{tab:circo}. In the table, we can observe that in all metrics, \ours achieves the best performances among the zero-shot CIR methods. This result supports the effectiveness of \ours and \ourdataset in real-world CIR tasks.

\textbf{GeneCIS.}
Finally, we report detailed GeneCIS in \cref{tab:genecis}. In the table, \ours shows the best performance in the average recall. Our method especially outperforms the other methods in ``Change Attribute'', ``Focus Object'' and ``Change Object''. On the other hand, \ours is less effective than Pic2Word and SEARLE in ``Focus attribute''. We presume that it is because the instruction distribution of ``Focus Attribute'' differs a lot from the instruction of \ourdataset. Among other CIR methods in the \ourdataset-trained group, \ours shows the best performances.

\section{More qualitative examples}
\label{subsec:more_examples_supp}

\begin{figure*}[t]
    \centering
    \includegraphics[width=\linewidth]{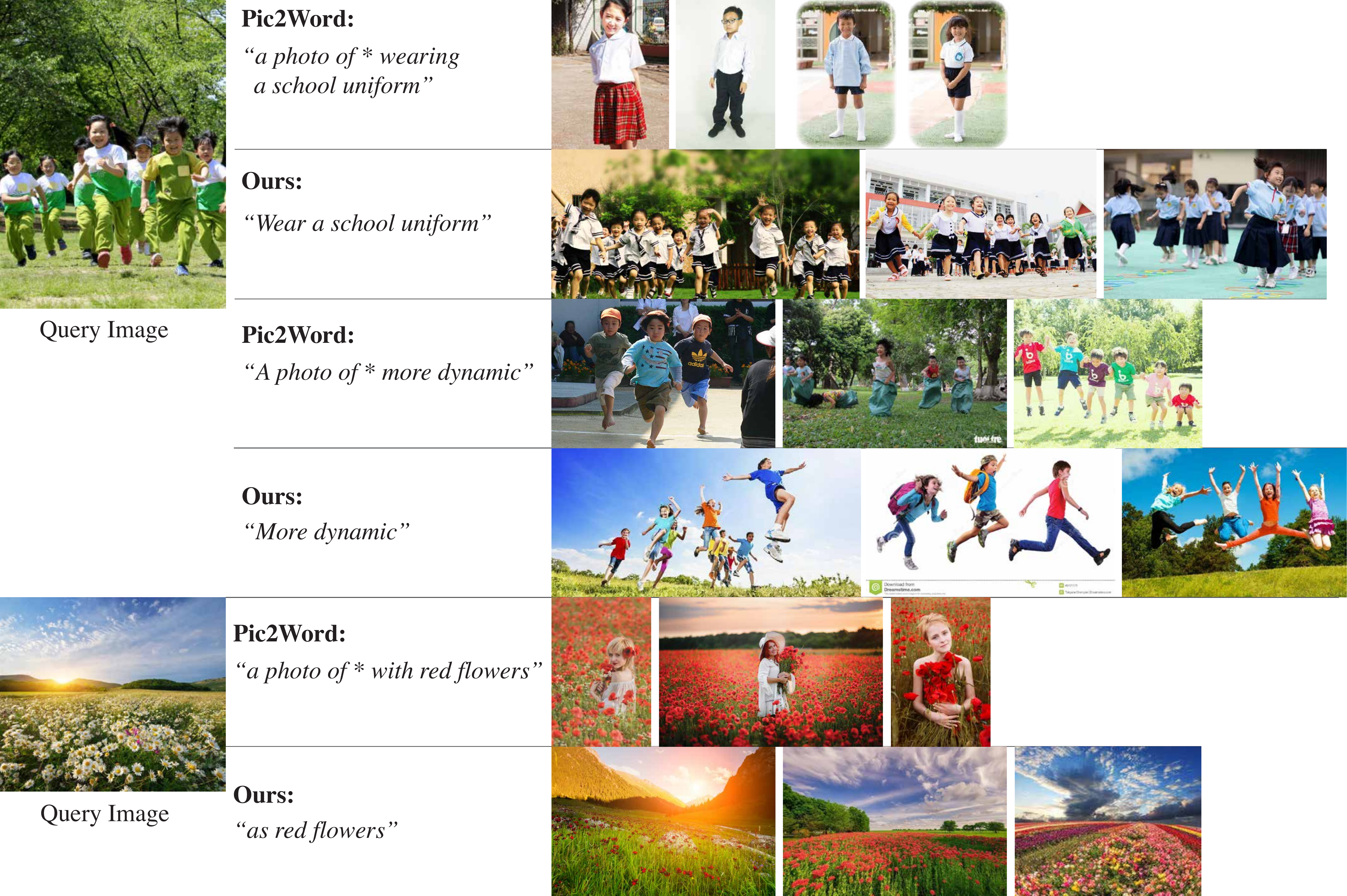}
    \vspace{-1em}
    \caption{\small {\bf More qualitative comparison of zero-shot CIR for Pic2Word and CompoDiff.}}
    \label{fig:ours_vs_pic2word_more_supp}
\end{figure*}

\paragraph{Open world zero-shot CIR comparisons with Pic2Word.} We illustrate further comparisons with Pic2Word in \cref{fig:ours_vs_pic2word_more_supp}. Here, we can draw the same conclusions as in the main text: Pic2Word often cannot understand images or instructions (\eg, ignores the ``crowdedness'' of the images, or retrieves irrelevant images such as images with a woman in the last example). All retrieved results in our paper were obtained using Pic2Word trained on the LAION 2B dataset.

\paragraph{More versatile CIR examples on LAION.}
\label{subsec:unclip_supp}

\begin{figure}[t]
    \centering
    \includegraphics[width=\linewidth]{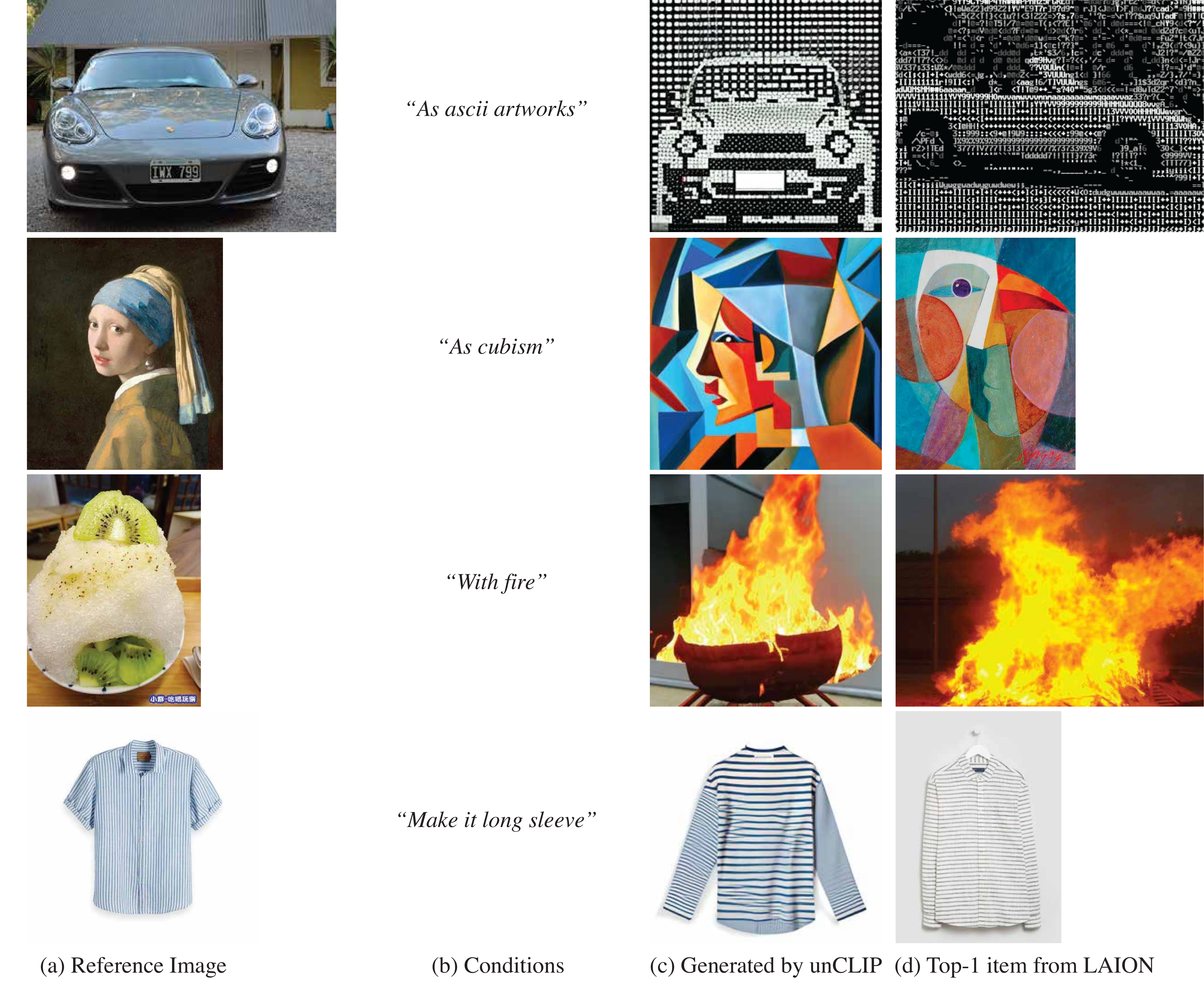}
    \caption{\small {\bf Generated vs. retrieved images by \ours.} Using the transformed image feature by \ours, Generated images using unCLIP and top-1 retrieved image from LAION.}
    \label{fig:more_examples_supp_0}
\end{figure}

\begin{figure}[t]
    \centering
    \includegraphics[width=\linewidth]{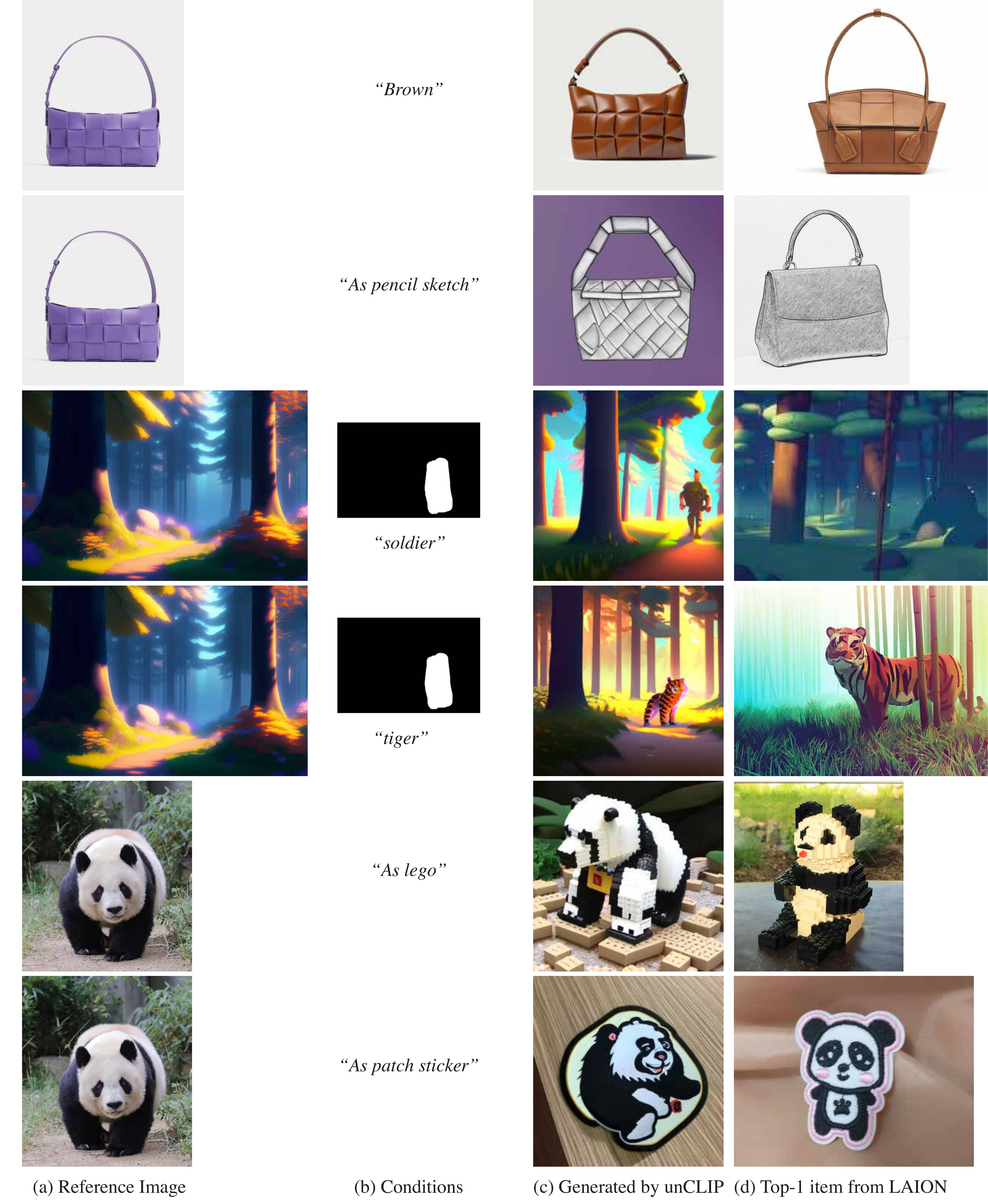}
    \caption{\small {\bf Generated vs. retrieved images by \ours (Continue).}}
    \label{fig:more_examples_supp_1}
\end{figure}

\begin{figure}[t]
    \centering
    \includegraphics[width=\linewidth]{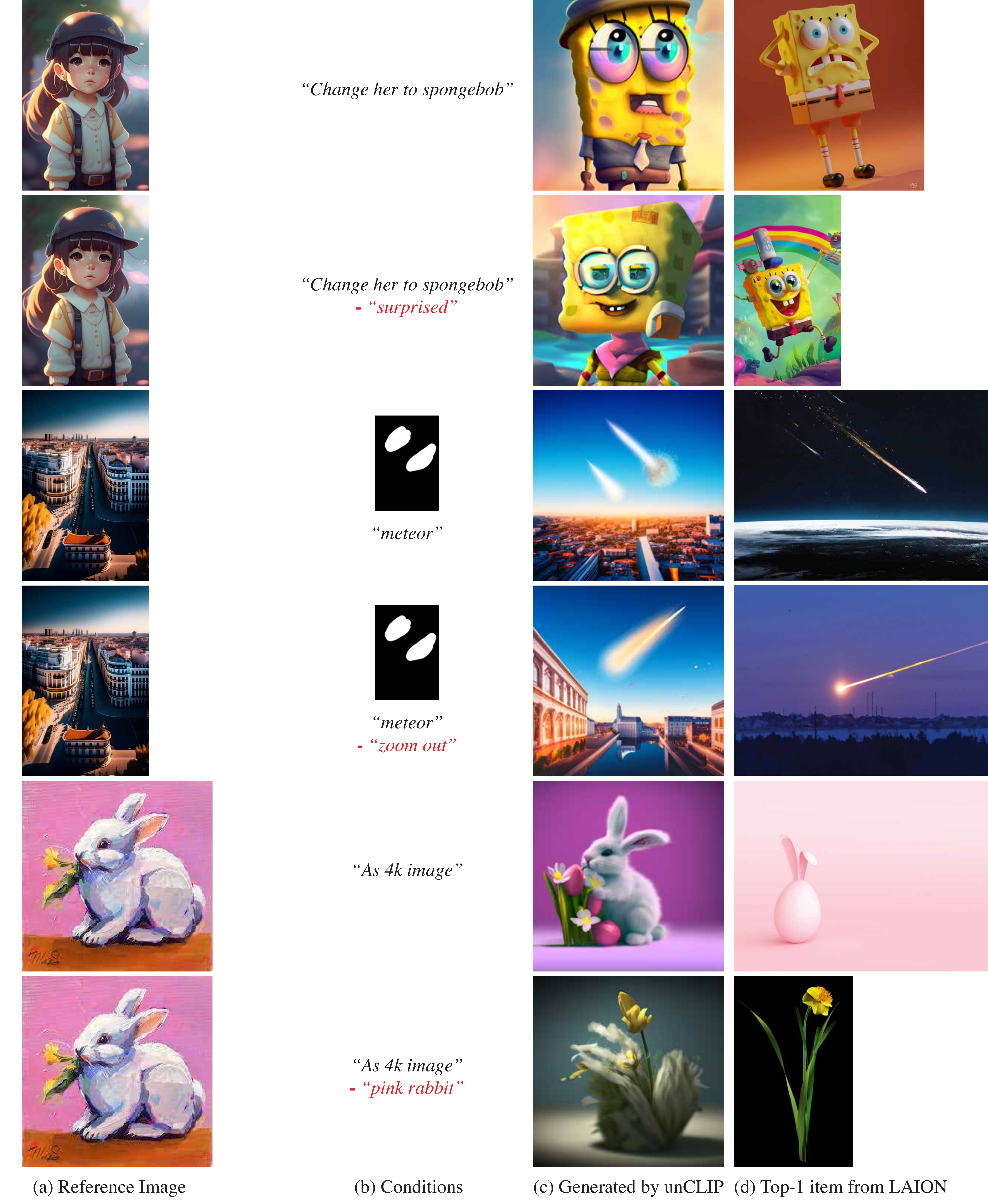}
    \caption{\small {\bf Generated vs. retrieved images by \ours (Continue).}}
    \label{fig:more_examples_supp_2}
\end{figure}

\begin{figure}[t]
    \centering
    \includegraphics[width=\linewidth]{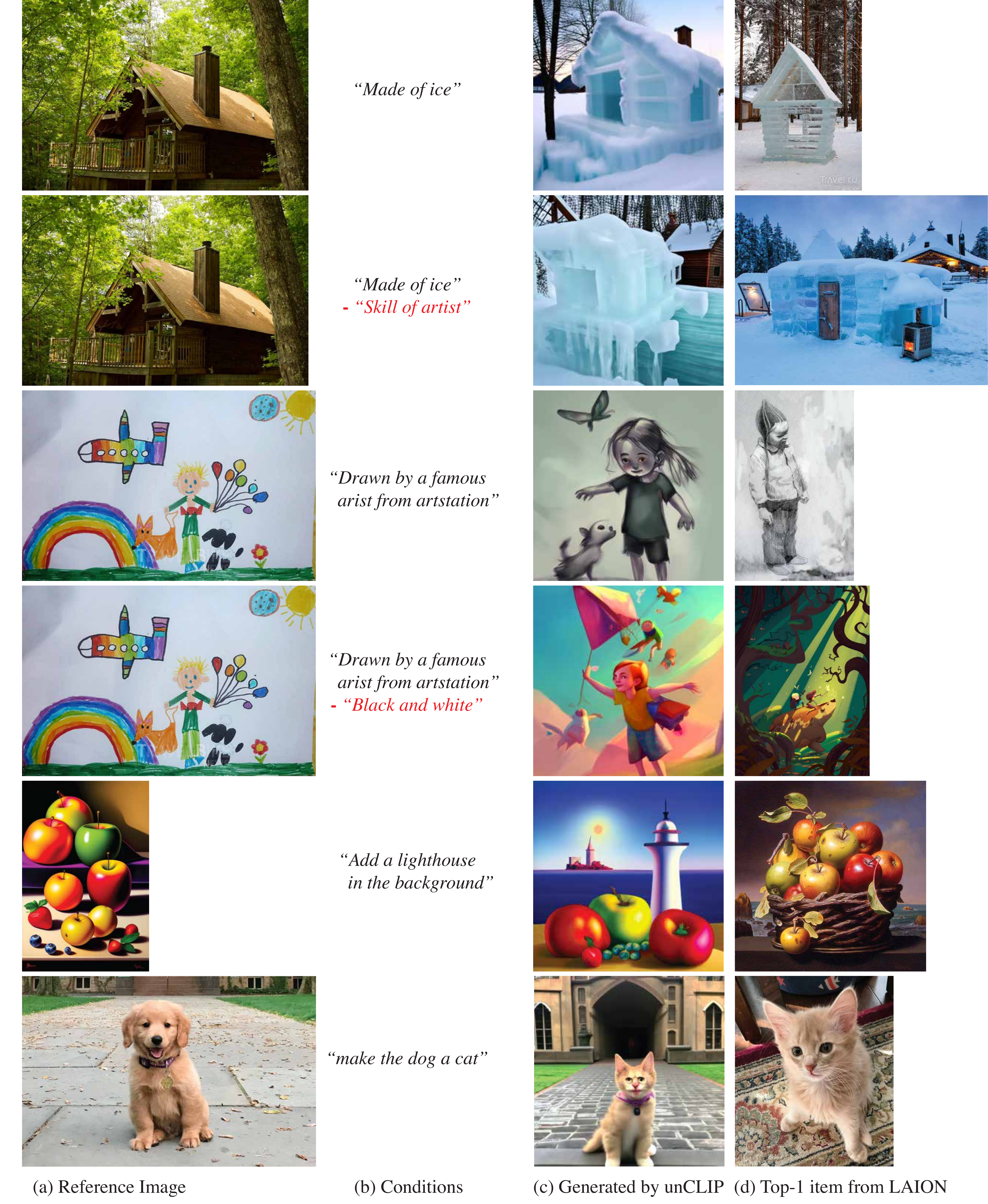}
    \caption{\small {\bf Generated vs. retrieved images by \ours (Continue).}}
    \label{fig:more_examples_supp_3}
\end{figure}

We illustrate more qualitative examples of the composed features by \ours, such as generated images by unCLIP retrieval results, in \cref{fig:more_examples_supp_0}, \cref{fig:more_examples_supp_1}, \cref{fig:more_examples_supp_2}, and \cref{fig:more_examples_supp_3}.

\clearpage

\clearpage

\end{document}

%% file: math_commands.tex
\usepackage{amsmath,amsfonts,bm}

\def\eqref#1{equation~\ref{#1}}

\def\1{\bm{1}}

\DeclareMathAlphabet{\mathsfit}{\encodingdefault}{\sfdefault}{m}{sl}
\SetMathAlphabet{\mathsfit}{bold}{\encodingdefault}{\sfdefault}{bx}{n}

%% file: tables/keywords.tex
\begin{table}[t]
\newcommand{\srctab}{\texttt{\$\{source\}}\xspace}
\newcommand{\trgtab}{\texttt{\$\{target\}}\xspace}
\setlength{\tabcolsep}{2pt}
\scriptsize
\centering
\begin{tabular}{lll}
\toprule
\multicolumn{3}{c}{Templates to change \srctab to \trgtab} \\ \midrule
\textit{``replace \srctab with \trgtab''} & \textit{``substitute \trgtab for \srctab''} & \textit{``change \srctab to \trgtab''} \\
\textit{``\trgtab''} & \textit{``\srctab is removed and \trgtab takes its place''} & \textit{``alter \srctab to \trgtab''} \\
\textit{``apply \trgtab''} & \textit{``modify \srctab to become \trgtab''} & \textit{``swap \srctab for \trgtab''} \\
\textit{``convert \srctab to \trgtab''} & \textit{``customize \srctab to become \trgtab''} & \textit{``redesign \srctab as \trgtab''} \\
\textit{``replace \srctab with \trgtab''} & \textit{``change \srctab to match \trgtab''} & \textit{``turn \srctab into \trgtab''} \\
\textit{``update \srctab to \trgtab''} & \textit{``\trgtab is introduced after \srctab is removed''} & \textit{``adapt \srctab to fit \trgtab''} \\
\textit{``substitute \trgtab for \srctab''} & \textit{``\trgtab is added in place of \srctab''} & \textit{``choose \trgtab instead''} \\
\textit{``alter \srctab to match \trgtab''} & \textit{``\trgtab is introduced as the new option after} & \textit{``\trgtab is the new choice''} \\
\textit{``upgrade \srctab to \trgtab''} & \textit{``\srctab is removed and \trgtab is added''} & \textit{``\trgtab is the new selection''} \\
\textit{``amend \srctab to fit \trgtab''} & \textit{``\srctab is removed and \trgtab is introduced''} & \textit{``\trgtab is the new option''} \\
\textit{``opt for \trgtab''} & \textit{``\trgtab is added as a replacement for \srctab''} & \textit{``use \trgtab from now on''} \\
\textit{\srctab is removed''} & \textit{``\trgtab is the new option available''} & \textit{``remodel \srctab into \trgtab''} \\
\textit{``add \trgtab''} & \textit{``\trgtab is added after \srctab is removed''} & \textit{``revamp \srctab into \trgtab''} \\
\textit{``if it is \trgtab''} & \textit{``\trgtab is introduced after \srctab is retired''} & \textit{``exchange \srctab with \trgtab''} \\
\textit{``\trgtab is the updated option''} & \textit{``tweak \srctab to become \trgtab''} & \textit{``transform \srctab into \trgtab''} \\
\textit{``\trgtab is the updated choice''} & \textit{``\srctab is replaced with \trgtab''} & \textit{``\trgtab is the updated version''}\\
\bottomrule
\end{tabular}
\caption{\small {\bf The full 48 keyword converting templates.}}
\label{tab:example_templates_supp}
\vspace{-1em}
\end{table}

%% file: tables/main_table.tex
\begin{table*}[t]
\small
\setlength{\tabcolsep}{5pt}
\centering
\begin{tabular}{ll|cc|cc|ccc|c}
\toprule
&& \multicolumn{2}{c|}{Fashion IQ (Avg)} & \multicolumn{2}{c|}{CIRR} & \multicolumn{3}{c|}{CIRCO} & GeneCIS \\
Method & Arch & R@10 & R@50 & R@1 & $\text{R}_\text{s}$@1 & {\footnotesize mAP}{\scriptsize @5} & {\footnotesize mAP}{\scriptsize @10} & {\footnotesize mAP}{\scriptsize @25} & R@1 \\
\midrule
CLIP + IP2P$^\dagger$ & ViT-L & 7.01 & 12.33 & 4.07 & 6.11 & 1.83 & 2.10 & 2.37 & 2.44 \\ \midrule
\multicolumn{10}{c}{Previous zero-shot methods (without \ourdataset)} \\ \midrule
Pic2Word$^\dagger$ & ViT-L & 24.70 & 43.70 & 23.90 & 53.76 & 8.72 & 9.51 & 10.65 & 11.16 \\
SEARLE-OTI$^\dagger$ & ViT-L & 27.51 & 47.90 & \underline{24.87} & {53.80} & 10.18 & 11.03 & 12.72 & - \\
SEARLE$^\dagger$ & ViT-L & 25.56 & 46.23 & 24.24 & {53.76} & {11.68} & {12.73} & {14.33} & {12.31} \\ \midrule
\multicolumn{10}{c}{Zero-shot results with the models trained with \ourdataset} \\ \midrule
ARTEMIS & RN50 & 33.24 & 47.99 & 12.75 & 21.95 & 9.35 & 11.41 & 13.01 & 13.52 \\
Combiner & RN50 & 34.30 & {49.38} & 12.82 & 24.12 & 9.77 & 12.08 & 13.58 & {14.93} \\
\ours & RN50 & 35.62 & 48.45 & 18.02 & 57.16 & 12.01 & 13.28 & 15.41 & 14.65 \\
\ours & ViT-L & {36.02} & 48.64 & 18.24 & {57.42} & {12.55} & {13.36} & \underline{15.83} & 14.88 \\
\ours & ViT-L \& T5-XL & \underline{37.36} & \underline{50.85} & {19.37} & \underline{59.13} & \underline{12.31} & \underline{13.51} & 15.67 & \underline{15.11} \\
\ours & ViT-G & \textbf{39.02} & \textbf{51.71} & \textbf{26.71} & \textbf{64.54} & \textbf{15.33} & \textbf{17.71} & \textbf{19.45} & \textbf{15.48} \\
\bottomrule
\end{tabular}
\caption{\small \textbf{Zero-shot CIR comparisons.} $\dagger$ denotes the results by the official model weight, otherwise, models are trained on \ourdataset and LAION-2B (ARTEMIS and Combiner are trained solely on \ourdataset, while \ours is trained on both). $\ddagger$ refers to the use of both the CLIP textual encoder and T5-XL as the encoder for the text condition. The full results for each dataset are shown in \cref{subsec:full_experiment_results_supp}.}
\label{tab:main}
\vspace{-.5em}
\end{table*}

%% file: tables/ablation_results_per_steps.tex
\begin{table}[h]
\small
\centering
\setlength{\tabcolsep}{3pt}
\resizebox{\textwidth}{!}{
\begin{tabular}{c|cccccccc|cccc}
\toprule
Step & 1 & 2 & 3 & 4 & 5 & 10 & 50 & 100 & Pic2Word & SEARLE & ARTEMIS & Combiner \\
\midrule
Avg(R@10, R@50) & 21.52 & 29.22 & 37.63 & \best{41.08} & \best{42.17} & \best{42.33}& 42.45 & 42.65 & 34.20 & 35.90 & 40.61 & 41.84 \\
Time (sec) & 0.02 & 0.05 & 0.07 & \best{0.09} & \best{0.12} & \best{0.23} & 1.08 & 2.02 & 0.02 & 0.02 & 0.005 & 0.006 \\
GPU Memory (bs 1) & \multicolumn{8}{c|}{2.60 GB} & {0.97 GB} & {1.32 GB} & {0.26 GB} & {0.54 GB} \\
GPU Memory (bs 256) & \multicolumn{8}{c|}{5.80 GB} & {2.99 GB} & {3.24 GB} & {2.82 GB} & {1.95 GB} \\
\bottomrule
\end{tabular}
}
\caption{\small \textbf{Performances vs. inference time by varying the number of denoising steps.} Numbers are measured on the FashionIQ validation split. \ours (ViT-L) in \cref{tab:main} is equivalent to 10 steps, but using 5 steps is a practical alternative. The inference time was measured on a single A100 GPU with a batch size of 1. GPU memories are the pick-allocated memory measured on batch sizes of 1 and 64, respectively. Note that ARTEMIS cannot support efficient batch operation for inference because its forward path needs triplet information, not an image-instruction pair; here, we report ARTEMIS information with training triplets.}
\label{tab:denoising_steps}
\end{table}

%% file: tables/i2t_retrieval.tex
\begin{table*}[t]
\small
\centering
\setlength{\tabcolsep}{3pt}
\begin{tabular}{lccc|ccc|ccc|c}
\toprule
 & \multicolumn{3}{c|}{COCO 5k} & \multicolumn{3}{c|}{CxC} & \multicolumn{3}{c|}{ECCV Caption} & Throughput \\
Method & R@1 & R@5 & R@10 & R@1 & R@5 & R@10 & mAP@R & R-P & R@1 & images/sec\\
\midrule
\multicolumn{11}{c}{Image to text retrieval} \\
\midrule
Prior \citep{kakaobrain2022karlo-v1-alpha} & 32.04 & 56.84 & 67.68 & 33.76 & 60.50 & 71.32 & 14.19 & 23.37 & 46.47& 497.28 \\
Prior-like Stage1 & 34.32 & 58.40 & 69.52 & 35.01 & 62.35 & 74.21 & 16.35 & 25.20 & 49.01 & 497.28 \\
Ours (Stage 1 only) & 35.13 & 59.46 & 70.26 & 35.30 & 62.62 & 74.02 & 16.01 & 25.44 & 49.64 & 1475.52 \\
Ours (Stage 1 + Stage2) & 33.20 & 58.00 & 68.94 & 34.78 & 61.68 & 72.96 & 15.07 & 24.39 & 47.03 & 1473.92\\
\midrule
\multicolumn{11}{c}{Text to image retrieval} \\
\midrule
Prior \citep{kakaobrain2022karlo-v1-alpha} & 17.05 & 34.25 & 43.16 & 18.62 & 37.43 & 47.07 & 18.10 & 26.46 & 46.40 & 497.28 \\
Prior-like Stage1 & 22.62 & 38.31 & 48.11 & 21.42 & 41.42 & 51.79 & 20.70 & 29.80 & 51.50 & 497.28 \\
Ours (Stage 1 only) & 22.47 & 39.18 & 49.08 & 22.51 & 42.40 & 52.77 & 21.46 & 30.30 & 53.75 & 1475.52 \\
Ours (Stage 1 + Stage2) & 20.00 & 38.63 & 48.25 & 21.57 & 41.71 & 51.99 & 20.82 & 29.84 & 51.65 & 1473.92\\
\bottomrule
\end{tabular}
\caption{\small \textbf{Comparisons of design choices for handling textual embeddings on cross-modal retrieval benchmarks.} Throughput was measured on 1 A100 GPU with a batch size of 32. For all metrics, higher is better. Here, we use the community version of the prior model by \citet{kakaobrain2022karlo-v1-alpha} because the official prior model \citep{dall-e2} is not public.}
\label{tab:i2t_retrieval}
\vspace{-.5em}
\end{table*}

%% file: tables/ablation_dataset_scale.tex
\begin{wraptable}{r}{0.47\textwidth}
\small
\centering
\vspace{-1.5em}
\setlength{\tabcolsep}{4pt}
\begin{tabular}{lccccc}
\toprule
& IP2P(1M) & 1M & 5M & 10M & 18.8M \\ \midrule
\multicolumn{6}{c}{FashionIQ Avg(R@10, R@50)} \\ \midrule
ARTEMIS & 26.03 & 27.44 & 36.17 & 41.35 & 40.62 \\
Combiner & 29.83 & 29.64 & 35.23 & 41.81 & 41.84 \\
\ours & 27.24 & 31.91 & 38.11 & 42.41 & 42.33 \\
\midrule
\multicolumn{6}{c}{CIRR Avg(R@1, R$_s$@1)} \\ \midrule
ARTEMIS & 14.91 & 15.12 & 15.84 & 17.56 & 17.35 \\
Combiner & 16.50 & 16.88 & 17.21 & 18.77 & 18.47 \\
\ours & 27.42 & 28.32 & 31.50 & 37.25 & 37.83 \\
\bottomrule
\end{tabular}
\vspace{-.5em}
\caption{\small \textbf{Impact of dataset scale.}
IP2P denotes the public 1M synthetic dataset by \citep{brooks2022instructpix2pix}.}
\label{tab:dataset_scale}
\vspace{-2em}
\end{wraptable}

%% file: tables/ablation_stage2.tex
\begin{wraptable}{r}{0.3\textwidth}
\small
\centering
\setlength{\tabcolsep}{4pt}
\vspace{-2em}
\begin{tabular}{lcc}
\toprule
& \textit{only} \cref{eq:stage2-triplet} & Proposed \\
\midrule
FIQ & 41.15 & \textbf{42.33} \\
CIRR & 35.25 & \textbf{37.83} \\
\bottomrule
\end{tabular}
\caption{\small \textbf{Stage 2 ablation.} 
Compared by FashionIQ Avg(R@10, R@50) and CIRR Avg(R@1 $\text{R}_s$@1).}
\label{tab:ablation_stage2}
\vspace{-2em}
\end{wraptable}

%% file: tables/parameters.tex
\begin{table}[h!]
\small
\centering
\begin{tabular}{lrrr}
\toprule
& Stage1 & Stage2 & Fine-tuning \\
\midrule
Diffusion steps & 1000 & 1000 & 1000 \\
Noise schedule & cosine & cosine & cosine \\
Sampling steps & 10 & 10 & 10 \\
Sampling variance method & DDIM & DDIM & DDIM \\ \midrule
Dropout & 0.1 & 0.1 & 0.1 \\
Weight decay & 6.0e-2 & 6.0e-2 & 6.0e-2 \\
Batch size & 4096 & 2048 & 2048 \\
Iterations & 1M & 200K & 50K \\
Learning rate & 1e-4 & 1e-5 & 1e-5 \\
Optimizer & AdamW & AdamW & AdamW \\
EMA decay & 0.9999 & 0.9999 & 0.9999 \\ \midrule
Input tokens & $z_i^{(t)}$, t & $z_i^{(t)}$, t & $z_i^{(t)}$, t \\
Conditions & $z_{c_T}$ & $z_{c_T}$, $z_{i_R}$, $z_{c_M}$ & $z_{c_T}$, $z_{i_R}$ \\
Training dataset & LAION-2B & LAION-2B, \ourdataset & FashionIQ or CIRR trainset \\
Image encoder & CLIP-L/14 & CLIP-L/14 & CLIP-L/14 \\
Text encoder & CLIP-L/14 & CLIP-L/14 & CLIP-L/14 \\
Denoiser depth & 12 & 12 & 12 \\
Denoiser heads & 16 & 16 & 16 \\
Denoiser head channels & 64 & 64 & 64 \\
\bottomrule
\end{tabular}
\caption{\small {\bf Hyperparameters.} A model trained by Stage 1 and Stage 2 is equivalent to ``Zero-shot'' in the main table. A ``supervised model'' is the same as the fine-tuned version.}
\label{tab:parameters}
\vspace{-.5em}
\end{table}

%% file: tables/fashion_iq_main.tex
\begin{table*}[t]
\small
\setlength{\tabcolsep}{5pt}
\centering
\begin{tabular}{lcc|cc|cc|ccc}
\toprule
 & \multicolumn{2}{c|}{Shirt} & \multicolumn{2}{c|}{Dress} & \multicolumn{2}{c|}{Toptee} & \multicolumn{3}{c}{Average} \\
Method & R@10 & R@50 & R@10 & R@50 & R@10 & R@50 & R@10 & R@50 & Avg. \\
\midrule
\multicolumn{10}{c}{Previous zero-shot methods (without \ourdataset)} \\ \midrule
CLIP + InstructPix2Pix & 7.24 & 12.71 & 6.31 & 10.42 & 7.49 & 13.85 & 7.01 & 12.33 & 9.67 \\
Pic2Word & 26.20 & 43.60 & 20.00 & 40.20 & 27.90 & 47.40 & 24.70 & 43.70 & 34.20 \\
SEARLE-XL-OTI & 30.37 & 47.49 & 21.57 & 44.47 & 30.90 & 51.76 & 27.61 & 47.90 \\
SEARLE-XL & 26.89 & 45.58 & 20.48 & 43.13 & 29.32 & 49.97 & 25.56 & 46.23 \\ \midrule
\multicolumn{10}{c}{Zero-shot results with the models trained with \ourdataset} \\ \midrule
ARTEMIS$^\dagger$ & 30.70 & 50.43 & 33.52 & 46.54 & 35.49 & 47.01 & 33.24 & 47.99 & 40.62 \\
CLIP4Cir$^\dagger$ & 32.32 & 51.65 & \underline{34.92} & \underline{48.38} & 35.65 & 48.10 & 34.30 & 49.38 & 41.84 \\
\ours (ViT-L)$^\dagger$ & 37.69 & 49.08 & 32.24 & 46.27 & {38.12} & {50.57} & 36.02 & 48.64 & {42.33} \\
\ours$^\ddagger$ (ViT-L)$^\dagger$ & \underline{38.10} & \underline{52.48} & 33.91 & 47.85 & \underline{40.07} & \underline{52.22} & \underline{37.36} & \underline{50.85} & \underline{44.11} \\
\ours (ViT-G)$^\dagger$ & \textbf{41.31} & \textbf{55.17} & \textbf{37.78} & \textbf{49.10} & \textbf{44.26} & \textbf{56.41} & \textbf{39.02} & \textbf{51.71} & \textbf{46.85} \\
\midrule
\multicolumn{10}{c}{Supervised} \\
\midrule
JVSM & 12.00 & 27.10 & 10.70 & 25.90 & 13.00 & 26.90 & 11.90 & 26.60 & 19.25 \\
CIRPLANT w/ OSCAR & 17.53 & 38.81 & 17.45 & 40.41 & 21.64 & 45.38 & 18.87 & 41.53 & 30.20 \\
TRACE w/ BERT & 20.80 & 40.80 & 22.70 & 44.91 & 24.22 & 49.80 & 22.57 & 46.19 & 34.38 \\
VAL w/ GloVe & 22.38 & 44.15 & 22.53 & 44.00 & 27.53 & 51.68 & 24.15 & 46.61 & 35.38 \\
MAAF & 21.30 & 44.20 & 23.80 & 48.60 & 27.90 & 53.60 & 24.30 & 48.80 & 36.55 \\
ARTEMIS & 21.78 & 43.64 & 27.16 & 52.40 & 29.20 & 54.83 & 26.05 & 50.29 & 38.17 \\
CurlingNet & 21.45 & 44.56 & 26.15 & 53.24 & 30.12 & 55.23 & 25.90 & 51.01 & 38.46 \\
CoSMo & 24.90 & 49.18 & 25.64 & 50.30 & 29.21 & 57.46 & 26.58 & 52.31 & 39.45 \\
RTIC-GCN w/ GloVe & 23.79 & 47.25 & 29.15 & 54.04 & 31.61 & 57.98 & 28.18 & 53.09 & 40.64 \\
DCNet & 23.95 & 47.30 & 28.95 & 56.07 & 30.44 & 58.29 & 27.78 & 53.89 & 40.84 \\
AACL & 24.82 & 48.85 & 29.89 & 55.85 & 30.88 & 56.85 & 28.53 & 53.85 & 41.19 \\
SAC w/ BERT & 28.02 & 51.86 & 26.52 & 51.01 & 32.70 & 61.23 & 29.08 & 54.70 & 41.89 \\
MUR & 30.60 & 57.46 & 31.54 & 58.29 & 37.37 & 68.41 & 33.17 & 61.39 & 47.28 \\
Combiner& 39.99 & \underline{60.45} & 33.81 & \underline{59.40} & 41.41 & \underline{65.37} & 38.32 & \underline{61.74} & \underline{50.03} \\ \midrule
\multicolumn{10}{c}{Pre-training with \ourdataset \& Fine-tuning with FashionIQ} \\ \midrule
ARTEMIS$^\dagger$ & 32.17 & 53.32 & 34.80 & 48.10 & 36.58 & 47.63 & 34.52 & 49.68 & 42.10 \\
Combiner$^\dagger$ & 37.21 & \textbf{60.71} & \textbf{42.75} & \textbf{60.50} & {42.98} & \textbf{65.49} & \textbf{40.98} & \textbf{62.23} & \textbf{51.61} \\
\ours (ViT-L)$^\dagger$ & \underline{40.88} & 53.06 & 35.53 & 49.56 & 41.15 & 54.12 & 39.05 & 52.34 & 46.31 \\
\ours (ViT-L + T5-XL)$^\dagger$ & {40.65} & {57.14} & 36.87 & {57.39} & \underline{43.93} & 61.17 & \underline{40.48} & 58.57 & 49.53 \\
\ours (ViT-G)$^\dagger$ & \textbf{41.68} & {56.02} & \underline{38.39} & {51.03} & \textbf{45.70} & {57.32} & {39.81} & {51.90} & {47.73} \\
\bottomrule
\end{tabular}
\caption{\small \textbf{Comparisons on FashionIQ.} The ``Zero-shot'' scenario performs CIR using a model not trained on the FashionIQ dataset, while models are trained on FashionIQ for the ``Supervised'' scenario. The last group denotes that a model is trained on \ourdataset.
$\dagger$ denotes that the models are newly trained by us.
}
\label{tab:fashioniq}
\vspace{-.5em}
\end{table*}

%% file: tables/cirr_main.tex
\begin{table*}[t]
\small
\setlength{\tabcolsep}{5pt}
\centering
\begin{tabular}{lcccc|ccc|c}
\toprule
Method & R@1 & R@5 & R@10 & R@50 & $\text{R}_\text{s}$@1 & $\text{R}_\text{s}$@2 & $\text{R}_\text{s}$@3 & Avg(R@1, $\text{R}_\text{s}$@1) \\
\midrule
\multicolumn{9}{c}{Previous zero-shot methods (without \ourdataset)} \\ \midrule
CLIP + InstructPix2Pix & 4.07 & 8.41 & 11.2 & 15.38 & 6.11 & 10.05 & 13.33 & 5.09 \\
Pic2Word & \underline{23.90} & 51.70 & 65.30 & 87.80 & 53.76 & 74.46 & 87.08 & 38.83 \\ 
SEARLE-XL-OTI & 24.87 & 52.31 & 66.29 & 88.58 & 53.80 & 74.31 & 86.94 & 39.34 \\
SEARLE-XL & 24.24 & 52.48 & 66.29 & 88.84 & 53.76 & 75.01 & 88.19 & 39.00 \\ \midrule
\multicolumn{9}{c}{Zero-shot results with the models trained with \ourdataset} \\ \midrule
ARTEMIS$^\dagger$ & 12.75 & 33.84 & 47.75 & 80.20 & 21.95 & 43.88 & 62.06 & 17.35 \\
CLIP4Cir$^\dagger$ & 12.82 & 36.83 & 48.19 & 81.91 & {24.12} & {46.47} & {63.07} & {18.47} \\
\ours (ViT-L)$^\dagger$ & 18.24 & {53.14} & {70.82} & {90.25} & {57.42} & {77.10} & {87.90} & {37.83} \\
\ours$^\ddagger$ (ViT-L)$^\dagger$ & {19.37} & \underline{53.81} & \underline{72.02} & \underline{90.85} & \underline{59.13} & \underline{78.81} & \underline{89.33} & \underline{39.25} \\
\ours (ViT-G)$^\dagger$ & \textbf{26.71} & \textbf{55.14} & \textbf{74.52} & \textbf{92.01} & \textbf{64.54} & \textbf{82.39} & \textbf{91.81} & \textbf{45.63}\\
\midrule
\multicolumn{9}{c}{Supervised} \\
\midrule
TIRG & 14.61 & 48.37 & 64.08 & 90.03 & 22.67 & 44.97 & 65.14 & 18.64 \\
TIRG + LastConv & 11.04 & 35.68 & 51.27 & 83.29 & 23.82 & 45.65 & 64.55 & 17.43 \\
MAAF & 10.31 & 33.03 & 48.30 & 80.06 & 21.05 & 41.81 & 61.60 & 15.68 \\
MAAF + BERT & 10.12 & 33.10 & 48.01 & 80.57 & 22.04 & 42.41 & 62.14 & 16.08 \\
MAAF-IT & 9.90 & 32.86 & 48.83 & 80.27 & 21.17 & 42.04 & 60.91 & 15.54 \\
MAAF-RP & 10.22 & 33.32 & 48.68 & 81.84 & 21.41 & 42.17 & 61.60 & 15.82 \\
CIRPLANT & 15.18 & 43.36 & 60.48 & 87.64 & 33.81 & 56.99 & 75.40 & 24.50 \\
CIRPLANT w/ OSCAR & 19.55 & 52.55 & 68.39 & 92.38 & 39.20 & 63.03 & 79.49 & 29.38 \\
ARTEMIS & 16.96 & 46.10 & 61.31 & 87.73 & 39.99 & 62.20 & 75.67 & 28.48 \\
Combiner & \underline{38.53} & \underline{69.98} & \underline{81.86} & \underline{95.93} & \underline{68.19} & \underline{85.64} & \underline{94.17} & \underline{53.36} \\
\midrule
\multicolumn{9}{c}{Pre-training with \ourdataset \& Fine-tuning with FashionIQ} \\ \midrule
ARTEMIS$^\dagger$ & 18.85 & 51.44 & 68.01 & 91.93 & 38.85 & 62.00 & 77.68 & 28.85 \\
Combiner$^\dagger$  & \textbf{39.99} & \textbf{73.63} & \textbf{86.77} & \textbf{96.55} & \textbf{68.41} & \textbf{86.12} & \textbf{94.80} & \textbf{54.20} \\
\ours (ViT-L)$^\dagger$ & 21.30 & 55.01 & 72.62 & 91.49 & 58.82 & 77.60 & 88.37 & 40.06 \\
\ours (ViT-L + T5-XL)$^\dagger$ & 22.35 & 54.36 & 73.41 & 91.77 & 62.55 & 81.44 & 90.21 & 42.45 \\
\ours (ViT-G)$^\dagger$ & {32.39} & {57.61} & {77.25} & {94.61} & 67.88 & 85.29 & 94.07 & 50.14 \\
\bottomrule
\end{tabular}
\caption{\small \textbf{Compasions on CIRR Test set.} Details are the same as \cref{tab:fashioniq}.}
\label{tab:cirr}
\vspace{-.5em}
\end{table*}

%% file: tables/circo_main.tex
\begin{table*}[t]
\small
\setlength{\tabcolsep}{5pt}
\centering
\begin{tabular}{lcccc}
\toprule
Method & mAP@5 & mAP@10 & mAP@25 & mAP@50 \\
\midrule
\multicolumn{5}{c}{Previous zero-shot methods (without \ourdataset)} \\ \midrule
CLIP + InstructPix2Pix & 1.83 & 2.10 & 2.37 & 2.44 \\
Pic2Word & 8.72 & 9.51 & 10.65 & 11.29 \\
SEARLE-XL & {11.68} & {12.73} & {14.33} & {15.12} \\ \midrule
\multicolumn{5}{c}{Zero-shot results with the models trained with \ourdataset} \\ \midrule
ARTEMIS & 9.35 & 11.41 & 13.01 & 14.13 \\
Combiner & 9.77 & 12.08 & 13.58 & 14.11 \\
\ours (ViT-L) & \underline{12.55} & 13.36 & \underline{15.83} & \underline{16.43} \\
\ours (ViT-L + T5-XL) & {12.31} & \underline{13.51} & {15.67} & {16.15} \\
\ours (ViT-G) & \textbf{15.33} & \textbf{17.71} & \textbf{19.45} & \textbf{21.01} \\
\bottomrule
\end{tabular}
\caption{\small \textbf{Compasions on CIRCO Test set.} Details are the same as \cref{tab:fashioniq}.}
\label{tab:circo}
\vspace{-.5em}
\end{table*}

%% file: tables/genecis_main.tex
\begin{table*}[t]
\scriptsize
\setlength{\tabcolsep}{5pt}
\centering
\begin{tabular}{lccc|ccc|ccc|ccc|c}
\toprule
Method & \multicolumn{3}{c|}{Focus Attribute} & \multicolumn{3}{c|}{Change Attribute} & \multicolumn{3}{c|}{Focus Object} & \multicolumn{3}{c|}{Change Object} & Avg\\
& R@1 & R@2 & R@3 & R@1 & R@2 & R@3 & R@1 & R@2 & R@3 & R@1 & R@2 & R@3 & R@1\\
\midrule
\multicolumn{14}{c}{Previous zero-shot methods (without \ourdataset)} \\ \midrule
Pic2Word & \underline{15.65} & \underline{28.16} & \underline{38.65} & 13.87 & 24.67 & 33.05 & 8.42 & 18.01 & \underline{25.77} & 6.68 & 15.05 & 24.03 & 11.16 \\
SEARLE-XL & \textbf{17.00} & \textbf{29.65} & \textbf{40.70} & 16.38 & 25.28 & 34.14 & 7.96 & \underline{16.94} & 25.61 & 7.91 & 16.79 & 24.80 & 12.31 \\ \midrule
\multicolumn{14}{c}{Zero-shot results with the models trained with \ourdataset} \\ \midrule
ARTEMIS & 11.76 & 21.97 & 25.44 & 15.41 & 25.14 & 33.10 & 8.08 & 16.77 & 24.70 & \underline{18.84} & 30.53 & 39.98 & 13.52 \\
Combiner & 14.11 & 24.08 & 34.12 & 18.39 & 28.22 & 37.13 & 8.49 & 16.70 & 25.21 & 18.72 & 30.92 & 40.11 & 14.93 \\
\ours (ViT-L) & 13.50 & 24.32 & 36.11 & 19.20 & 28.64 & \underline{37.20} & 8.11 & 16.39 & 25.08 & 18.71 & 31.69 & 40.55 & 14.88 \\
\ours$^\ddagger$ (ViT-L) & {13.01} & {25.01} & {36.75} & \textbf{19.88} & \underline{28.64} & {37.18} & \underline{8.60} & {16.28} & \textbf{25.85} & \textbf{18.94} & \textbf{31.80} & \textbf{40.58} & \underline{15.11} \\
\ours (ViT-G) & {14.32} & {26.70} & {38.41} & \underline{19.72} & \textbf{28.78} & \textbf{37.39} & \textbf{9.18} & \textbf{19.11} & \underline{25.77} & {18.71} & \underline{31.71} & \underline{40.22} & \textbf{15.48} \\ \bottomrule
\end{tabular}
\caption{\small \textbf{Compasions on GeneCIS Test set.} Details are the same as \cref{tab:fashioniq}.}
\label{tab:genecis}
\vspace{-.5em}
\end{table*}